\documentclass{article} %
\usepackage{iclr2025_conference,times}

\usepackage{amsmath,amsfonts,bm}

\def\eqref#1{equation~\ref{#1}}

\def\1{\bm{1}}

\def\eps{{\epsilon}}

\DeclareMathAlphabet{\mathsfit}{\encodingdefault}{\sfdefault}{m}{sl}
\SetMathAlphabet{\mathsfit}{bold}{\encodingdefault}{\sfdefault}{bx}{n}

\DeclareMathOperator*{\argmax}{arg\,max}
\DeclareMathOperator*{\argmin}{arg\,min}

\usepackage[utf8]{inputenc} %
\usepackage[T1]{fontenc}    %
\usepackage{hyperref}       %
\usepackage{url}            %
\usepackage{booktabs}       %
\usepackage{amsfonts}       %
\usepackage{nicefrac}       %
\usepackage{microtype}      %
\usepackage{xcolor}        
\usepackage{wrapfig}
\usepackage{adjustbox}
\usepackage{amsmath}
\usepackage{graphicx}
\usepackage{mystyle}
\usepackage{amsmath}
\usepackage{xcolor}  
\usepackage{apxproof}
\usepackage{tikz} %
\usepackage{hyperref}
\ifdefined\nohyperref\else\ifdefined\hypersetup
  \definecolor{mydarkblue}{rgb}{0.1,0.1,0.6}
  \hypersetup{ %
    pdftitle={},
    pdfauthor={},
    pdfsubject={},
    pdfkeywords={},
    pdfborder=0 0 0,
    pdfpagemode=UseNone,
    colorlinks=true,
    linkcolor=mydarkblue,
    citecolor=mydarkblue,
    filecolor=mydarkblue,
    urlcolor=mydarkblue,
    pdfview=FitH}

\colorlet{michaels}{violet}

\newcommand{\mic}[1]{{#1}}

\definecolor{ForestGreen}{cmyk}{0.91,0,0.88,0.12}

\usepackage{enumitem}

\newcommand{\tmax}{T_{\mathrm{max}}}
\title{{Towards Understanding the Universality of Transformers for Next-Token Prediction}}

\author{{Michaël E. Sander \& Gabriel Peyré} \\
Ecole Normale Supérieure, CNRS \\
Paris, France \\
\texttt{michael.sander@polytechnique.org}, \texttt{gabriel.peyre@ens.fr} \\
}

\iclrfinalcopy %
\begin{document}

\maketitle

\begin{abstract}

Causal Transformers are trained to predict the next token for a given context. While it is widely accepted that self-attention is crucial for encoding the causal structure of sequences, the precise underlying mechanism behind this in-context autoregressive learning ability remains unclear. 
In this paper, we take a step towards understanding this phenomenon by studying the approximation ability of Transformers for next-token prediction. Specifically, we explore the capacity of causal Transformers to predict the next token $x_{t+1}$ given an autoregressive sequence $(x_1, \dots, x_t)$ as a prompt, where $ x_{t+1} = f(x_t) $, and $ f $ is a context-dependent function that varies with each sequence.
On the theoretical side, we focus on specific instances, namely when $ f $ is linear or when $ (x_t)_{t \geq 1} $ is periodic. We explicitly construct a Transformer (with linear, exponential, or softmax attention) that learns the mapping $f$ in-context through a causal kernel descent method.
The causal kernel descent method we propose provably estimates $x_{t+1} $ based solely on past and current observations $ (x_1, \dots, x_t) $, with connections to the Kaczmarz algorithm in Hilbert spaces.
We present experimental results that validate our theoretical findings and suggest their applicability to more general mappings $f$.
\end{abstract}

\section{Introduction}
The transformative impact of deep learning on artificial intelligence has led to increasingly powerful architectures, with Transformers \citep{vaswani2017attention} at the forefront. These models have in particular revolutionized natural language processing (NLP) \citep{devlin2018bert}, and now serve as the foundation for large-scale language models, such as GPT \citep{radford2018improving, brown2020language}, significantly advancing artificial intelligence's capabilities in both understanding and generating human language, setting new benchmarks across various tasks.

Most recent large language models \citep{hoffmann2022training, team2023gemini, jiang2023mistral, dubey2024llama} are causal Transformers pretrained to predict the most likely next token $x_{t+1}$ from a finite vocabulary given a context $ x_{1:t} = (x_1, \cdots, x_t) $. These models excel at such tasks and beyond, demonstrating what is known as in-context learning: after training, they show remarkable few-shot learning capabilities, inferring patterns from just a few examples within the context \citep{brown2020language}.
Recent studies suggest that in-context learning capabilities emerge from the Transformer performing optimization on an inner objective during its forward pass, where the attention matrix plays a crucial role \citep{von2023transformers, mahankali2023one, ahn2023transformers, zhang2023trained, kim2024transformers}. In particular, \citet{chengtransformers} demonstrate that trained Transformers can perform in-context learning through kernel ridge regression. However, the question of why causal Transformers excel at general autoregressive prediction remains open.

\begin{figure}[H]
\centering
\includegraphics[width=1\columnwidth]{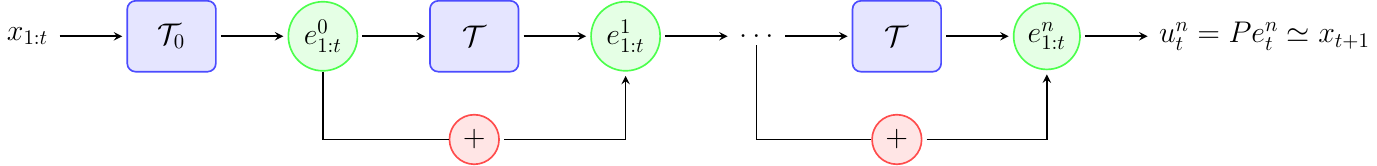} 
\vspace{-1em}
\caption{\small{Illustration of the method proposed in this paper. Given a sequence $x_{1:t}$, a first layer $\mathcal{T}_0$ computes augmented tokens $e^0_{1:t}$. Next, a stack of $n$ identical Transformer layers $\mathcal{T}$ with residual connections iteratively update the tokens $e^k_{1:t}$, following the causal kernel descent method introduced in Section \ref{sec:causal_kernel_descent}. 
For autoregressive sequences presented in \mic{Assumption} \ref{def:seq}, and under specific instances outlined in \mic{Assumption} \ref{def:instances}, projecting $e^n_t$ with a projector $P$ yields an estimate $u^n_t$ of $x_{t+1}$ as $n$ and $t$ approach $+\infty$, as stated in Theorem \ref{thm:main}.}}
\label{fig:intro}
\end{figure}

In this paper, we propose a kernel interpretation for the autoregressive setting. We introduce a framework to rigorously analyze the expressivity of deep Transformers in next-token prediction. Specifically, we consider sequences generated according to $ x_{t+1} = f(x_t) $, with $ f $ a context-dependent function in a vector-valued Reproducing Kernel Hilbert Space (RKHS) associated with a positive semi-definite kernel $k$. This generalizes the works of \citet{vonoswald2023uncovering, sander2024transformers}. 
Within this framework, we explore how successive attention layers solve a causal kernel least square regression problem to predict the next token accurately, as described in Figure \ref{fig:intro}.
{More precisely, we make the following contributions:}
\vspace{-.5em}
\begin{itemize}[left=1pt]
    \item In Section \ref{sec:transformer_implem}, we formalize the sequence generation model that serves as the foundation for our theoretical results.
   We then present our main result in Theorem \ref{thm:main}, demonstrating that there exists a Transformer model, with an explicit construction, that, given the first $t $ tokens $x_{1:t}$, accurately predicts the next token $x_{t+1} $ as $t $ tends to $+\infty $, for specific instances, namely when $f $ is linear or when the sequence $(x_t)_{t \geq 1} $ is periodic.
    \item In Section \ref{sec:causal_kernel_descent}, we present the methodology underlying the proof of Theorem \ref{thm:main}. We introduce a family of causal kernel descent methods that build an estimate $ u^\star_t $ of $ x_{t+1} $, based only on past and current observations $ x_{1:t} $. This approach modifies a least squares gradient descent method to account for causality while preserving the parallelization benefits of the Transformer architecture. In Theorems \ref{thm:linear_linear}, \ref{thm:stationary}, and \ref{thm:periodic}, we prove that for the specific cases considered, $ u^{\star}_t - x_{t+1} $ converges to 0 as $ t \to +\infty $, drawing connections to the Kaczmarz algorithm \citep{kaczmarz1937approximate} in Hilbert spaces. Finally, Proposition \ref{prop:discrete} shows that the causal kernel descent methods can be implemented with a Transformer model, as illustrated in Figure \ref{fig:intro}. 
    \item In Section \ref{sec:experiments}, we first present experimental results that validate our theoretical findings and extend them to a more general class of mappings $f$ beyond those studied in Sections \ref{sec:transformer_implem} and \ref{sec:causal_kernel_descent}. We then empirically show that the Transformer models constructed in Theorem \ref{thm:main} can be successfully fine-tuned to obtain faster convergence of the estimate with the sequence length $t$.
\end{itemize}

\section{Background and related work}

\paragraph{Transformers.}

Transformers \citep{vaswani2017attention} process sequences of tokens $(x_1, \cdots, x_T)$ or arbitrarily length $T$. After embedding the sequence into a new sequence $(e_1, \cdots, e_T)$, the Transformer consists of a series of blocks with residual connections \citep{he2016deep}. Each block comprises two primary components: a multi-head self-attention mechanism and a feedforward multi-layer perceptron, with the latter operating independently on each token. 
We will almost always disregard the feedforward layer in this study (except when building augmented tokens), as is common in theoretical analyses of in-context learning \citep{mahankali2023one, ahn2023transformers, zhang2023trained}.
In contrast, the multi-head self-attention mechanism involves pairwise interaction between the tokens. This module consists in applying multiple self-attention operations in parallel, parametrized by a set of weight matrices $(W^h_Q, W^h_K, W^h_V)_{1\leq h \leq H}$, where $H$ denotes the number of attention heads \citep{vaswani2017attention, michel2019sixteen}.
The output of the multi-head self-attention mechanism is given by:
\begin{equation}\label{eq:MHSA}
e_t \leftarrow e_t + {\Tt(e_{1:t})}, \quad \text{with} \quad {\Tt(e_{1:t})} \eqdef \sum_{h=1}^{H} \sum_{s=1}^{t} \mathcal{A}^h_{t,s} W^h_V e_{s},
\end{equation}
where $\mathcal{A}^{h}$, the attention matrix, determines the attention weights between tokens and is typically defined as:
$
\mathcal{A}^{h}_{t,:}=\Nn(\langle W^h_Q e_{t} , W^h_Ke_{:}\rangle ),
$
with $\langle \cdot , \cdot \rangle$ representing a dot product and $\Nn$ being a normalization function.
The standard choice is to consider $\Nn=\text{softmax}$, i.e. $$\mathcal{A}^h_{t,s} = {{e^{\langle W^h_Q e_{t} , W^h_Ke_{s}\rangle}}/{\sum_{\tau=1}^t e ^{\langle W^h_Q e_{t} , W^h_Ke_{\tau}\rangle} }}.$$  One can also consider the unormalized attention when $\Nn = \text{exp}$. Another approach is to consider $\Nn = \text{id}$, which corresponds to what is known (despite being non-linear) as linear attention \citep{katharopoulos2020transformers}, enabling faster inference.
One significant advantage of \eqref{eq:MHSA} is that the updates on the $e_t$'s can be computed in parallel during training, leveraging modern hardware for faster computations.

\paragraph{Expressivity.}  The universal approximation properties of encoder-only Transformers are well established. \citet{yun2019transformers, nath2024transformers, furuya2024transformers} demonstrate that Transformers can approximate permutation-equivariant functions. A more constructive approach, though applicable to a narrower class of functions, is proposed by \citet{wang2024understandingexpressivepowermechanisms}.
When it comes to decoder-only models, the expressivity of Transformers for next-token prediction is not yet fully understood, though a popular recent line of works studies the in-context learning ability of Transformers.
\vspace{-1em}
\paragraph{In-context learning.} A major property of Transformers is that they adapt their computations given the context. In particular, given a context $(x_1, g(x_1), \cdots, x_n)$, a trained large Transformer can infer the next output $g(x_n)$ without parameter updates. Many recent studies have contributed to understanding this phenomenon. The seminal work of \citet{von2023transformers} considers functions $g$ of the form $g(x) = w^\top x$ for some $w$ and construct a linear Transformer for which the forward pass is equivalent to a single step of gradient descent on a mean squared error loss. 
Theoretical guarantees are provided by \citet{mahankali2023one, ahn2023transformers, zhang2023trained}, showing a trained one-layer linear Transformer implements one step of (preconditioned) gradient descent. 
Other works study the $\text{softmax}$ attention without accounting for training dynamics \citep{garg2022can, akyurek2022learning, li2023transformers}. 
Of particular interest to us is the recent work of \citet{chengtransformers} which shows that there exists a simple parameter configuration of non-linear Transformers such that they implement gradient descent in the function space with respect to the RKHS metric induced by the attention kernel. In this work, we take a step further by considering a more general next-token prediction task and propose a causal kernel descent method that can be implemented by a Transformer to solve it. 
For this, we extend the autoregressive in-context learning setting introduced by \citet{sander2024transformers}, where tokens are generated according to an autoregressive process of order $1$: $x_{t+1} = f(x_t)$, where $f(x) = W x $ for a context-dependant parameter $W$, varying with each sequence. The autoregressive in-context learning ability is described as the model's capacity to decompose its prediction into two steps: first, estimating $ W $ through an in-context mapping, and then applying a straightforward prediction function, which is either equal to or closely related to $ x \mapsto W x$. However, \citet{sander2024transformers} focus solely on linear Transformers, whereas our study encapsulates linear, exponential, and softmax Transformers. 
\mic{Our key contribution is showing how such Transformers can implement an optimization algorithm--termed causal kernel descent--within their forward pass. This can be interpreted as a mesa-optimization mechanism \citep{vonoswald2023uncovering}.} 

\paragraph{Neural Ordinary Differential Equations.}
\vspace{-1em}
Neural ODEs \citep{e2017dynamical, chen2018neural} are a class of implicit models where a neural network $\mathcal{F}$ parameterizes a vector field in an ordinary differential equation (ODE), as follows: 
\begin{equation}\label{eq:node}
    \frac{de}{d\tau}(\tau) = \mathcal{F}(e(\tau)),
\end{equation}
where $\tau$ denotes the continuous depth of the network. For a given input $e(0)$, a neural ODE outputs a deep representation $e^{\star}$, which is the solution (when it exists) of \eqref{eq:node} with initial condition $e(0)$, at some finite or infinite time horizon. Neural ODEs can be viewed as a continuous-depth analog of deep Residual Networks \citep{he2016deep}, where the latter corresponds to an Euler discretization for solving \eqref{eq:node} \citep{marion2023implicit}. 
As such, neural ODEs are widely employed to better understand the theoretical properties of Residual Networks.
In the context of Transformers, the neural ODE framework has emerged as a valuable tool for studying attention-based models, such as the impact of the choice of normalization function $\Nn$ on model behavior \citep{sander2022sinkformers} and the emergence of clusters \citep{geshkovski2024emergence}. However, these studies do not address the autoregressive setting, which we tackle in this work.
\vspace{-.5em}
\section{Transformers for Autoregressive In-Context Learning}\label{sec:transformer_implem}
\vspace{-.5em}
\paragraph{Notations.} \mic{Throughout the paper,} we denote $d$ the dimension of $x_t$, $\| .\|$ the $\ell_2$ norm, $O(d)$ the orthogonal manifold, $S^{d-1}$ the unit sphere in $\RR^d$ and $^*$ the adjoint.

In this section, we describe the types of autoregressive sequences we consider and present our main results, which show that we can explicitly construct Transformer models that approximate the next token in the sequence.
\vspace{-.5em}
\subsection{Proposed Framework}

\paragraph{Reproducing Kernel Hilbert Space.}

For $ k : \mathbb{R}^d \times \mathbb{R}^d \to \mathbb{R} $ a positive definite kernel, we define $\mathcal{H}$ as the vector-valued Reproducing Kernel Hilbert Space (RKHS) associated with the feature map $\phi : \mathbb{R}^d \to \mathcal{H}$, where $\phi(x) = k(x, \cdot)$. 
Therefore, for any function $f \in \Hh$, there exists a linear map $W : \Hh \to \RR^d$ such that $f(x) = W \phi(x)$, where, denoting $W = (w_1, \cdots ,w_d)$ as $d$ vectors of $\Hh$, $\|W\|_{\Hh} \eqdef \sum_{i=1}^d \|w_i \|^2_{\Hh}$ is finite.

We consider autoregressive sequences of order $1$, which we formalize in the following \mic{assumption}.
\begin{asp}[Autoregressive sequences.]\label{def:seq}
\mic{We consider sequences of order $1$, defined as follows:}
    \begin{itemize}[left=0pt]
    \item \textbf{Initial State}: The sequence starts with some $x_1 \in S^{d-1}$.
    \item \textbf{Hidden Variable}: We suppose there is a hidden variable $f \in \Hh$ such that the subsequent states are generated autoregressively as
    $
    x_{t+1} = f(x_t)
    $ for $t \geq 1$.
\end{itemize}
\end{asp}

Note that although we focus on first-order recursions here, higher-order recursions can be considered by embedding tokens in a higher-dimensional space.
\mic{Indeed, for recursions of the form $x_{t+1} = g(x_t, \cdots, x_{t-\tau})$ for some context mapping $g$, defining $y_t \eqdef (x_t, \cdots, x_{t-\tau}) \in \RR^{(\tau+1)d}$, one has $y_{t+1} = (x_{t+1}, \cdots, x_{t+1-\tau}) = (g(x_t, \cdots, x_{t-\tau}), \cdots, x_{t+1-\tau}) $ which only depends on $y_t$. We thus have $y_{t+1} = f(y_t)$ for some mapping $f$. Therefore as long as the recursion memory is finite, our approach can be generalized.}
The formulation in \mic{Assumption} \ref{def:instances} differs from the classical in-context learning setup, where sequences consist of input-output pairs $ (x_i, y_i) $, that are often supposed to be iid. We argue that modeling sequences with $ x_{t+1} = f(x_t) $ better reflects the nature of real-world sequences on which causal Transformers, such as large language models \mic{(LLMs)}, are trained. \mic{Indeed, our proposed model incorporates autoregressive relationships, aligning more closely with the data LLMs encounter during training.}
However, similarly to in-context learning, in order to accurately predict the next token $x_{t+1}$ given the previous states $x_{1:t}$ as inputs, a Transformer would have to implicitly estimate the hidden map $f$. In this paper, we propose a general method to provide an estimate of $x_{t+1}$ given $x_{1:t}$. However, to prove that such an estimate can be built with a standard Transformer, and in order to prove our universality results in Theorem \ref{thm:main}, we consider specific choices for the kernel $k$ and the function $ f $, summarized in the following \mic{assumption}.

\begin{asp}[Specific considerations.]\label{def:instances} 
We define the following instances.
\begin{itemize}[left=1pt, label={}]
\item  \textbf{(1)} $ k(x,y) = k_{\text{id}}(x,y) \eqdef  \langle x , y \rangle$, $ f(x) = W x $ for some $W \in O(d)$.%
\item  \textbf{(2)}  $ k(x,y)=k_{\text{exp}}(x,y)\eqdef e^{\langle x , y \rangle} $ and $ f(x) = \Omega x $ for some $ \Omega \in O(d) $.
\item  \textbf{(3)} $ k=k_{\text{exp}} $, the sequence $(x_t)_{t\geq1}$ is periodic and $x_t \in S^{d-1}$ for all $t$.
\end{itemize}
\end{asp}

Note that for these specific kernel choices, because $\|x_t\| = 1$, one has $k(x_t, x_t) = k(x_1, x_1)$ for all $t$. \mic{This follows because $k(x_t, x_t)$ depends only on $\|x_t\| = \|x_1\|$.}

\paragraph{Augmented tokens.}

A crucial step in our construction is the building of augmented tokens for a sequence $ (x_1, \cdots, x_T) $. Indeed, in \citet{vonoswald2023uncovering} and \citet{sander2024transformers}, augmented tokens 
are used, explicitly encoding the relative positions of tokens $x_t$.
This approach is similar to in-context learning, where sequences often consist of pairs $ (x_i, y_i) $ to encode both input and output information. However, such a construction with attention-based models is non-trivial.  \citet{sander2024transformers} propose a method for constructing augmented tokens using general positional encoding. In this work, we provide a more detailed result, showing that augmented tokens can be computed with a one-layer Transformer that employs dot-product absolute positional encoding. We introduce a beginning-of-sequence token $x_0 \eqdef 0_d$ and consider the extended sequence $ x_{0:T} = (x_0, x_1, \cdots, x_T) $. 
In this work, we are going to consider the augmented tokens $e^0_t \in \RR^{4d+2}$ defined as 
$$ e^0_t \eqdef (x_{t-1} , 0, x_t, 1, x_t, 0_d) \text{ for }  t > 1, \text { and } e^0_1 \eqdef (0_{d}, 1, x_t, 1, 0_d, 0_d).$$ 
\mic{The intuition behind augmented tokens is that they enable the model to directly access two successive elements in the sequence. Importantly, while prior works, such as \citet{von2023transformers, vonoswald2023uncovering}, directly worked on augmented tokens, one of our contribution lies in showing how they can be computed using positional encoding-based attention. The augmented tokens incorporate standard beginning-of-sequence tokens, while the $1$'s correspond to concatenated positional encodings, which are technically required in our proofs.}
\mic{More precisely}, the following proposition shows that the $e^0_t$ can be approximated with a Transformer layer. 
\begin{prop}\label{prop:construction}
    There exists a sequence of one-layer and $2$-heads causal Transformer $ \mathcal{T}^n_0 $ with  $\Nn=\text{softmax}$ followed by a feedforward layer, such that $ \mathcal{T}_0(x_{0:t}) \eqdef  \lim_{n\to +\infty} \mathcal{T}^n_0(x_{0:t}) = e^0_t $.
\end{prop}

A proof of this result, along with the explicit construction of the corresponding model, is provided in Appendix \ref{proof:construction}. Here, $e^0_1$ should be interpreted as a new beginning-of-sequence token. The $0$'s and $1$'s in $e^0_t$ correspond to specific choices of positional encodings, which are crucial for handling the softmax normalization in our main Theorem \ref{thm:main}. 
From this point onward, we will consider the new augmented tokens $e^0_t$.

\subsection{Main result}
 
We now present our main theorem. Using the augmented tokens $e^0_t$, we demonstrate that a stack of Transformer layers can approximate the next token $x_{t+1}$ based solely on $e^0_{1:t}$ (and thus solely on $x_{1:t}$ as well) as $t$ increases.

\paragraph{Approximation with Transformers.}
We consider a model $\Mm^n$ composed of $n$ identical Transformer layers $\Tt$ with residual connections, followed by a projection. The model iterates from the beginning-of-sequence token $e_t^0$ as:
\begin{equation}\label{eq:discrete}
    \Mm^n(x_{1:t}) \eqdef Pe^n_t, \quad \text{where} \quad   e_t^{k+1} = e_t^k + \Tt(e_{1:t}^k), \quad 0 \leq k \leq n-1,
\end{equation}
with $P: \RR^{4d+2} \to \RR^d$ the projector selecting the last $d$ coordinates, which can be integrated in the form of a token-wise feedforward layer.

We have the following main theorem, which shows that for the specific instances we consider, there exist Transformer layers $\Tt$ such that $\lim_{t\to+\infty}\lim_{n\to +\infty}\Mm^n(x_{1:t}) - x_{t+1} = 0$.

\begin{thm}[\mic{On the expressivity} of Transformers for Next Token Prediction]
\label{thm:main}
For each instance in \mic{Assumption} \ref{def:instances}, there exists an attention-only, one-layer, two-head causal Transformer $\Tt$ with attention normalization $\Nn$ such that, for any autoregressive sequence $(x_{t})_{t\geq1}$ generated according to \mic{Assumption} \ref{def:seq}, $\Mm^n(x_{1:t})$ converges exponentially fast as $n$ goes to infinity. Furthermore, denoting $\Mm(x_{1:t}) \eqdef \lim_{n \to +\infty}\Mm^n(x_{1:t})$, one has $\lim_{t\to+\infty}(\Mm(x_{1:t})- x_{t+1})=0$. More specifically:
\begin{itemize}[left=0pt]
    \item For instance $(1)$, $\Nn = \text{id}$, and the convergence of $(\Mm(x_{1:t}) - x_{t+1})$ to $0$ 
    is exponentially fast in $t$ for almost all $x_1$ and $W$.
    \item For instance $(2)$, $\Nn = \text{exp}$ or $\Nn = \text{softmax}$.
    \item For instance $(3)$, $\Nn = \text{exp}$ or $\Nn = \text{softmax}$, and the convergence of $(\Mm(x_{1:t}) - x_{t+1})$ to $0$ is exponentially fast in $t$.
\end{itemize}
Finally, for any of the instances above, when $\Nn = \text{id}$ or $\Nn = \text{exp}$, we have $\Mm^n(x_{1:t}) = \Mm(x_{1:t})$ as long as $n \geq t$, so that $\lim_{n, t \to +\infty \text{, } n\geq t}(\Mm^n(x_{1:t})- x_{t+1})= 0$.
\end{thm}
We emphasize that the models $ \Tt $ correspond to explicit constructions. Specifically, we have:%
$$ 
W_Q^1, W_K^1 \in \RR^{d \times (4d + 2)}, \quad W_V^1, W_V^2 \in \RR^{(4d + 2) \times (4d + 2)}, \quad W_Q^2, W_K^2 \in \RR^{(d + 1) \times (4d + 2)}.
$$
\mic{More precisely, one has}

\adjustbox{max width=\textwidth}{
\mic{\begin{minipage}{0.48\textwidth}
    \scriptsize
    \begin{align*}
    W_Q^{(1)} &= W_K^{(1)} = [\mathbf{0}_{d \times d+1}, I_d, \mathbf{0}_{d \times 2d+1}], \\
    W_V^{(1)} &= -
    \begin{pmatrix}
        \mathbf{0}_{d+1 \times d+1} & \mathbf{0}_{d+1\times d+1} & \mathbf{0}_{d+1 \times d} & \mathbf{0}_{d+1 \times d} \\
        \mathbf{0}_{d+1 \times d+1} & \mathbf{0}_{d+1\times d+1} & \mathbf{0}_{d+1 \times d} & \mathbf{0}_{d+1 \times d} \\
        \mathbf{0}_{d \times d+1} & \mathbf{0}_{d\times d+1} & \mathbf{0}_{d \times d} & \mathbf{0}_{d \times d} \\
        \mathbf{0}_{d \times d+1} & \mathbf{0}_{d\times d+1} & \mathbf{0}_{d \times d} & I_d
    \end{pmatrix},
    \end{align*}
\end{minipage}
\hfill
\begin{minipage}{0.48\textwidth}
    \scriptsize
    \begin{align*}
    W_Q^{(2)} &= [\mathbf{0}_{d+1 \times d+1}, I_{d+1}, \mathbf{0}_{d+1 \times 2d}], \\
    W_K^{(2)} &= [I_{d+1}, \mathbf{0}_{d+1 \times d+1}, \mathbf{0}_{d+1 \times 2d}], \\
    W_V^{(2)} &= 
    \begin{pmatrix}
        \mathbf{0}_{d+1 \times d+1} & \mathbf{0}_{d+1 \times d+1} & \mathbf{0}_{d+1 \times d} & \mathbf{0}_{d+1 \times d} \\
        \mathbf{0}_{d+1 \times d+1} & \mathbf{0}_{d+1 \times d+1} & \mathbf{0}_{d+1 \times d} & \mathbf{0}_{d+1 \times d} \\
        \mathbf{0}_{d \times d+1} & \mathbf{0}_{d \times d+1} & \mathbf{0}_{d \times d} & \mathbf{0}_{d \times d} \\
        \mathbf{0}_{d \times d+1} & \mathbf{0}_{d \times d+1} & I_d & \mathbf{0}_{d \times d}
    \end{pmatrix}.
    \end{align*}
\end{minipage}}
}

The proof of Theorem \ref{thm:main}, which is in Appendix \ref{proof:main}, relies on demonstrating that the model $\Mm^n$ implements a causal kernel descent method, which we describe and analyze in Section \ref{sec:causal_kernel_descent}. 

\begin{rem}\label{rem:1}
    In Theorem \ref{thm:main}, we have $\|\Mm^n(x_{1:t}) - x_{t+1}\| \leq \eps_1(n,t) + \eps_2(t)$, with  $\eps_1(n,t) = \| \Mm^{n}(x_{1:t})- \Mm(x_{1:t}) \|$ and $\eps_2(t) = \| \Mm(x_{1:t})- x_{t+1} \|$. The proof of Theorem \ref{thm:main} reveals that $\lim_{n\to+\infty}\eps_1(n,t) = 0$ and $\lim_{t\to+\infty}\eps_2(t) = 0$. When $\Nn = \text{id}$ or $\Nn = \text{exp}$, $\eps_1(n,t) = 0$ if $n \geq t$, hence the last statement of the theorem. When $\Nn = \text{softmax}$, we conjecture that $\eps_1(n,t) \to 0$ as $n,t \to +\infty$ and $t/n \to 0$, which implies $\lim_{n,t \to +\infty \text{ and } t/n \to 0}(\Mm^n(x_{1:t}) - x_{t+1})=0$. We provide evidence for this conjecture in the last paragraph of Appendix \ref{proof:main}.
\end{rem}

The model $\Mm$ introduced in Theorem \ref{thm:main} also has a continuous-time interpretation, which we now formulate.
\vspace{-.5em}
\paragraph{Neural ODE Interpretation.}
The model $\Mm$ defined in Theorem \ref{thm:main} as the limit when $n$—the number of Transformer layers—goes to infinity can be interpreted as a continuous-time neural ODE \citep{chen2018neural}. Specifically, $\Mm$ satisfies
\begin{equation}\label{eq:anode}
\Mm(x_{1:t}) \eqdef \lim_{\tau \to +\infty}Pe_t(\tau), \quad \text{where} \quad \frac{de_t}{d\tau}(\tau) = \Tt(e_{1:t}(\tau)) 
\quad \text{with} \quad e_t(0) = \mathcal{T}_0(x_{0:t}).
\end{equation}

We now present the theory behind our constructions to guarantee consistent approximation of the next token $x_{t+1}$ as the sequence size increases.
\paragraph{Comparison with RNNs.}
\mic{Even though we consider autoregressive sequences, it is not straightforward that recurrent neural networks (RNNs) can effectively capture these models. This is because estimating $W$ in-context requires computations with long-range dependencies. Determining the optimal $W$ indeed requires inverting the data covariance matrix. Attention mechanisms inherently handle such procedure, which is proven in this work. We believe RNNs would require more layers to ``propagate'' such information. While each RNN layer is less computationally expensive than an attention layer, the overall cost might be similar. Investigating this is complex and beyond the scope of this article, which demonstrates how current Transformer-based architectures are particularly well-suited for in-context learning due to their global attention mechanism.}
\vspace{-.5em}
\section{Causal Kernel Descent}\label{sec:causal_kernel_descent}
In this section, we introduce a causal method to instantiate the iterations in \eqref{eq:discrete} and prove Theorem \ref{thm:main}. Specifically, we propose a causal kernel descent method that incorporates causality into standard gradient descent for least squares minimization.

\subsection{Causal Descent}
 
\paragraph{Non-Causal Descent.}

We consider a sequence $ x_{1:T} \eqdef (x_1, \cdots, x_T) \in \mathbb{R}^{T \times d} $. The goal is to solve the least squares minimization problem of minimizing $\sum_{s=1}^{T-1} \|f(x_s) - x_{s+1}\|^2$ with respect to $f \in \Hh$. Recall that for any $f \in \Hh$, there exists a linear map $W : \Hh \to \mathbb{R}^d$ such that $f(x) = W \phi(x)$ for all $x \in \mathbb{R}^d$. Thus, we consider the least squares optimization problem:
\vspace{-.5em}
\begin{equation}\label{eq:OLS}
    \min_W E(W) \eqdef \sum_{s=1}^{T-1} \|W \phi(x_s) - x_{s+1}\|^2 .
\end{equation}
We solve \eqref{eq:OLS} using gradient descent with step size $\frac{\eta}{2}$, starting from an initial $W^0$ and iterating for $0 \leq k \leq n-1$:
$$
W^{k+1}= W^k - \eta \sum_{s=1}^{T-1} (W^k \phi(x_s) - x_{s+1}) \phi(x_s)^* 
.
$$
We define a prediction variable $u^k_s \eqdef W^k \phi(x_s)$. By right-multiplying the gradient descent equation by $\phi(x_t)$, we obtain:
\vspace{-.5em}
\begin{equation}\label{eq:flow_prediction}
 u^{k+1}_t  =  u^{k}_t - \eta \sum_{s=1}^{T-1}k(x_t, x_s)  (u^k_s - x_{s+1}) 
,
\end{equation}
which corresponds to a least squares descent on the predictions. 
However, this descent is non-causal because the update of $u_t$ depends on $u_{1:T-1}$ and $x_{1:T}$, making it unsuitable for implementation in a causal Transformer. We now propose a causal formulation for \eqref{eq:flow_prediction}.
\vspace{-1em}
\paragraph{Causal Descent.}
Inspired by the descent in \eqref{eq:flow_prediction}, we propose a modified least squares descent that introduces causality, ensuring that each estimate of $x_{t+1}$ is based solely on past and current observations $x_{1:t}$.
To achieve this, we define the following \textit{causal kernel descent}, which incorporates both an unnormalized and a row-wise normalized framework. Starting from any initial $u^0_t$, the descent iterates for $0 \leq k \leq n-1$ as follows:
\begin{equation}\label{eq:discrete_causal}
u^{k+1}_{t} = u^k_{t} - \eta \sum_{s=1}^t A_{t,s} (u^k_{s}  - 1_{s<t}x_{s+1})
 \text{ with } A_{t,s}= 
\begin{cases} 
k(x_t, x_s), & \text{if } k = k_{\text{id}} \\
k(x_t, x_s) \text{ or } \frac{k(x_t, x_s)}{\sum_{\tau=1}^t k(x_t, x_{\tau})}, & \text{if } k = k_{\text{exp}}
\end{cases}.
\end{equation}
We denote $A$ the corresponding lower triangular matrix.
Note that this descent is causal in the sense that $u^n_t$ depends only on $x_{1:t}$. Note also that each iteration in \eqref{eq:discrete_causal} can be parallelized. 
For well-chosen step sizes $\eta$, the method in \eqref{eq:discrete_causal} converges. Specifically, we have the following proposition:
\begin{prop}\label{prop:discrete_causal}
For each causal kernel descent in \eqref{eq:discrete_causal}, there exists $\eta^{\star}$ and $u^{\star}_t$ such that, for all $0 < \eta < \eta^{\star}$, $u^n_t \to u^{\star}_t$ exponentially fast as $n$ goes to infinity. Specifically:
\begin{itemize}[left=0pt]
    \item When $A_{t,s} = k(x_t, x_s)$, then $\eta^{\star} = \frac{2}{k(x_1,x_1)}$. Moreover, when $\eta = \frac{1}{k(x_1,x_1)}$, $u^n_t = u^{\star}_t$ if $n \geq t$.
    \item When $A_{t,s} =  \frac{k(x_t, x_s)}{\sum_{\tau=1}^t k(x_t, x_\tau)}$, then $\eta^{\star} = 2$.
\end{itemize}
\end{prop}
Our proof, in Appendix \ref{proof:discrete_causal}, relies on the lower triangular property of the matrix $A$.

Our objective is to show that $ u^{\star}_t $ is ``close'' to $ x_{t+1} $ when $t$ is sufficiently big, demonstrating that the causal kernel descent accurately tracks the future states provided it has seen sufficiently long context.
We have the following result. 
\begin{prop}\label{prop:mu}
    Let $(\mu_t)_{t \geq 1}$ be the unique sequence of vectors satisfying
    \begin{equation}\label{eq:dual}
    \sum_{s=1}^t \mu_s k(x_s, x_t) = x_{t+1},   \quad \forall t \geq 1.
    \end{equation}
    Then for all $t \geq 1$, $x_{t+1} - u^{\star}_t = k(x_1, x_1) \mu_t.$
\end{prop}
For a proof, see Appendix \ref{proof:mu}.
From Proposition \ref{prop:mu}, we see that $\lim_{t \to +\infty}(u^{\star}_t - x_{t+1}) = 0$ is equivalent to having $\lim_{t \to +\infty} \mu_t = 0$. We provide a dual interpretation for $\mu$ in Appendix \ref{app:additional}.

\subsection{Convergence in the sequence length}
\begin{wrapfigure}{r}{0.4\textwidth}
\vspace{-1.5em}
\includegraphics[width=0.4\textwidth]{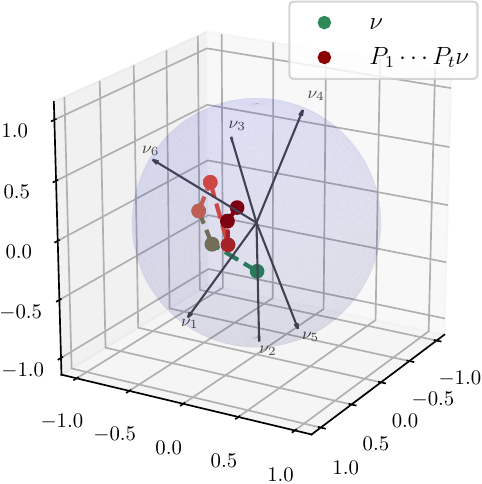}
\vspace{-2.em}
\caption{\small{For some random vectors $\nu$ (green) and $\nu_1, \cdots \nu_6$ in $S^2$, we display $P_6\nu$ (grey), $P_{5}P_6 \nu$ (orange), ... and $P_1 \cdots P_6 \nu$. }}
\vspace{-1.2em}
\label{fig:projectors}
\end{wrapfigure}
In this section, we prove that under the instances defined in \mic{Assumption} \ref{def:instances}, $\lim_{t \to +\infty}(u^{\star}_t - x_{t+1}) = 0$.
For any $ \mu $ satisfying \eqref{eq:dual}, we define the following map from $\Hh$ to $\RR^d$:
\vspace{-.5em}
\begin{equation}
    W_t \eqdef \sum_{s=1}^t \mu_s \phi(x_s)^*.
\end{equation}
By construction, we have for all $t$: $ W_t \phi(x_t) = x_{t+1} $. To prove that $ \lim_{t \to +\infty} \mu_t = 0 $ (and therefore $\lim_{t \to +\infty}(u^{\star}_t - x_{t+1}) = 0$), we will actually prove a stronger result, showing that for the instances in \mic{Assumption} \ref{def:instances} and in a specific limit sense, $ W_t \to W $.
For this, we define

\begin{equation}
    \nu_t \eqdef \frac{\phi(x_t)}{\sqrt{k(x_t, x_t)}}= \frac{\phi(x_t)}{\|\phi(x_t)\|_{\Hh}} \quad \text{and} \quad  P_t \eqdef \left(I - \nu_t \nu_t^* \right).
\end{equation}
Therefore, $P_t$ is the orthogonal projection onto the subspace orthogonal to $ \phi(x_t)$.
With these notations, we have the following proposition:
\begin{prop}[Estimate Update Recursion]\label{prop:diff}
For all $  t \geq 1 $, we have
$$
W_{t} - W =  - W P_1 P_2 \cdots P_t.
$$
\end{prop}
For a proof, see Appendix \ref{proof:diff}.  
Intuitively, the recursive relationship in Proposition \ref{prop:diff} shows how the difference between $ W_t $ and $ W $ is progressively reduced by successive orthogonal projections. \mic{Such a process is illustrated in Figure \ref{fig:projectors}, where each point corresponds to an iterate $\nu_t$, defined as $\nu_t = P_1 \cdots P_t \nu$.}
Note that without further assumptions, there is no guarantee that $(W_t - W)\phi(x)$ converges to $0$ for arbitrary $x$. However, for the specific instances considered in \mic{Assumption} \ref{def:instances}, we are able to establish convergence for vectors $x$ in a certain space.

We first focus on instance $(1)$, where we can derive convergence speed results.

\paragraph{Linear recursions and dot-product kernel.} In this section, we consider instance $(1)$ from \mic{Assumption} \ref{def:instances}.
Under these assumptions, we prove the following theorem:
\begin{thm}[$k = k_{\text{id}}$, linear recursions]\label{thm:linear_linear}
Under instance $(1)$ in \mic{Assumption} \ref{def:instances}, one has that $\lim_{t \to +\infty}(u^{\star}_t - x_{t+1}) = 0$. In addition, for almost all $W$ and $x_1$, $W_t \to W$ exponentially fast. 
\end{thm}
\begin{proofsketch}
We first show that $W_t - W = - ( W (I - x_1 x_1^\top))^t W^{-t+1}$. We then establish that for almost all $W$ and $x_1$, $\rho(W (I - x_1 x_1^\top)) < 1$, where $\rho$ denotes the spectral radius.
\end{proofsketch}
See Appendix \ref{proof:linear_linear} for a full proof. We now turn to the case where $k = k_{\text{exp}}$.

\paragraph{Linear recursions and exponential kernel.} In this paragraph, we consider instance $(2)$ in \mic{Assumption} \ref{def:instances}. %
We present the following theorem, with a complete proof provided in Appendix \ref{proof:stationary}.
\begin{thm}[$k=k_{\text{exp}}$, linear recursions]\label{thm:stationary}
Under instance $(2)$ in \mic{Assumption} \ref{def:instances}, for any $x \in \mathbb{R}^d$, $(W^*_t - W^*)x \to 0$. In particular, $\lim_{t \to +\infty}(u^{\star}_t - x_{t+1}) = 0$.
\end{thm}
\begin{proofsketch}
    We consider the subset of $ \Hh $ comprising functions
$
x \mapsto \sum_{s=1}^\tau a_s \nu_s, 
$
for $a_s \in \RR^d$ and $\tau \geq 1$. 
This subset is a pre-Hilbert space under the inner product inherited from $ \Hh $. By completing this space with respect to the induced norm from $\Hh$, we obtain a new Hilbert space $ \Hh' $. We show that $ P_t P_{t-1} \cdots P_1 \to 0 $ strongly in $\Hh'$ as $t \to +\infty$. 
For this, we observe that the convergence of $P_t P_{t-1} \cdots P_1$ is equivalent to the convergence of the Kaczmarz algorithm \citep{kaczmarz1937approximate}.
A sequence $ (\nu_s)_{s \geq 1} $ of unit vector for which $ P_t P_{t-1} \cdots P_1 \to 0 $ strongly as $t \to +\infty$ is referred to as \textit{effective}.
The specificities of our case are twofold: first, $ \nu_s  \in \mathcal{\Hh'} $, which is potentially of infinite dimension, and second, the vectors $ \nu_s $ follow an autoregressive relation.
We show that the sequence $(\nu_s)_{s \geq 1}$ is effective in $\Hh'$. 
Note that because $\Omega\in O(d)$, one has for any positive integers $t, s, r$ that $\langle \nu_{s + r}, \nu_{t + r} \rangle = \langle \nu_{s}, \nu_{t} \rangle$. Such sequence is called \textit{stationary}.
Bochner's theorem states that there exists a measure $\sigma$ on the unit circle $S^1$--called \textit{spectral measure}--such that, for all $t \geq 1$,
$$
 a_t \eqdef \langle \nu_{t+1}, \nu_1 \rangle = \int_{S^1} z^t \, d\sigma(z).
$$
We then use the following characterization from \citet{kwapien2001kaczmarz, RainisHaller2005}.
\vspace{-.5em}
\paragraph{Theorem}(Effectiveness of stationary sequences \citep{kwapien2001kaczmarz}).
\textit{A stationary sequence of unit vectors which is linearly dense in a Hilbert space is effective if and only if its spectral measure either coincides with the normalized Lebesgue measure or is singular with respect to the Lebesgue measure.}

This characterization combined with Fourier analysis results shows that the sequence $(\nu_s)_{s \geq 1}$ is effective. 
While this does not necessarily imply the strong convergence of $W_t - W = -W P_1 \cdots P_t$ to $0$, this is enough to prove that $\lim_{t\to+\infty}(u^{\star}_t - x_{t+1})=0$.
\end{proofsketch}

\paragraph{Periodic recursions.} In this paragraph, we turn to instance $(3)$ in \mic{Assumption} \ref{def:instances}, where $k = k_{\text{exp}}$ and the sequence $(x_t)_{t \geq 1}$ is assumed to be periodic. The motivation for considering periodic sequences is to better align with real-world next-token prediction tasks, where the vocabulary has a finite size and is frequently repeated in cyclical patterns. 
We have the following theorem (proof in Appendix \ref{proof:periodic}).
\begin{thm}[$k=k_{\text{exp}}$, periodic recursions.]\label{thm:periodic}
     Under instance $(3)$ in \mic{Assumption} \ref{def:instances}, one has that $\lim_{t\to+\infty}(u^{\star}_t - x_{t+1}) = 0$ exponentially fast.
\end{thm}

\subsection{Implementation with Transformers}

We can now use the previous results stating that the output of the causal kernel method $u^{\star}_t$ approaches $x_{t+1}$ as $t$ increases to construct the models presented in Section \ref{sec:transformer_implem}.

\paragraph{Expressing \eqref{eq:discrete_causal} with a Transformer.}
We have the following proposition, which, combined with  Theorems \ref{thm:linear_linear}, \ref{thm:stationary} and \ref{thm:periodic}, allow us to prove Theorem \ref{thm:main}.
\begin{prop}\label{prop:discrete}

For any $\eta >0$, for each configuration of \eqref{eq:discrete_causal}, there exists an attention-only, one-layer, two-head causal Transformer $\Tt$ with attention normalization $\Nn$ such that, for any autoregressive sequence $(x_{t})_{t\geq 1} $ generated according to \mic{Assumption} \ref{def:seq}, defining $e^{k}_t \eqdef (x_{t-1} , 0, x_t, 1, x_t, u^k_t) $ for $t>1$ and $e^k_1 \eqdef (0_{d}, 1, x_t, 1, 0_d, u^k_t)$, $e^{1:n}_{1:t}$ solves \eqref{eq:discrete} if and only if $u^{1:n}_{1:t}$ solves \eqref{eq:discrete_causal}. More specifically,
\begin{itemize}[left=0pt]
    \item When $A_{t,s} = k(x_t, x_s)$ and $k = k_{\text{id}}$, we have $\Nn = \text{id}$.
    \item When $A_{t,s} = k(x_t, x_s)$ and $k = k_{\text{exp}}$, we have $\Nn = \text{exp}$.
    \item When $A_{t,s} = \frac{k(x_t, x_s)}{\sum_{\tau=1}^tk(x_t, x_{\tau})}$ and $k = k_{exp}$, we have $\Nn = \text{softmax}$.
\end{itemize}
\end{prop}
Here again, we stress that the transformer layers $\Tt$ correspond to an explicit construction. See Appendix \ref{proof:discrete} for a constructive proof.
With Proposition \ref{prop:discrete}, we prove Theorem \ref{thm:main}, in Appendix \ref{proof:main}.

\begin{figure}[ht]
\centering
\includegraphics[width=1\columnwidth]{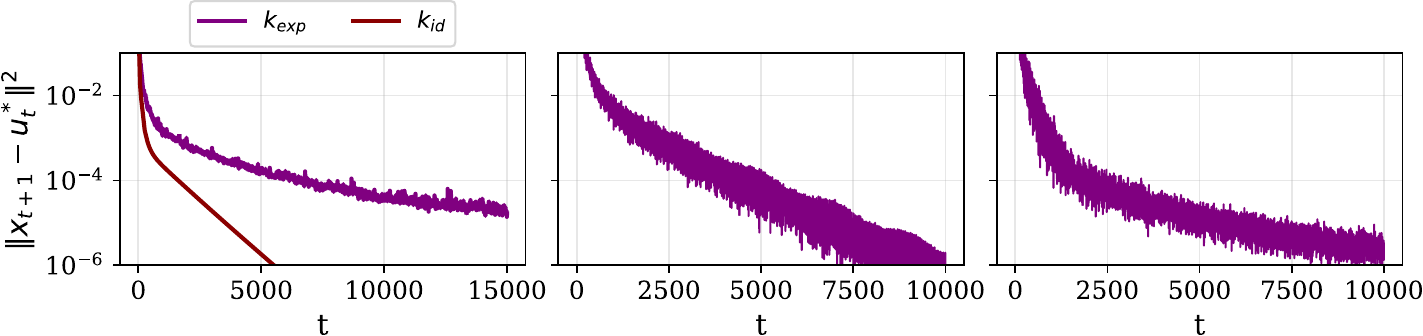} 
\caption{\small{\textbf{Evolution} of the squared error $\|u^{\star}_t - x_{t+1}\|^2$ with $t$ for different scenarios. The curves are averaged over five sequences $x_{1:t}$. Left: instances (1) and (2) (with random $W$ and $\Omega$), illustrating Theorems \ref{thm:linear_linear} and \ref{thm:stationary} ($d = 15$). Center: instance $(3)$, illustrating Theorem \ref{thm:periodic} ($d = 15$, the period $t_p$ is randomly sampled between $20$ and $40$, and a random sequence is repeated $t_p$ times). Right: instance (4) described in Section \ref{sec:experiments} ($d = 4$).}}\label{fig:illustration_theorems}
\end{figure}

\section{Experiments} \label{sec:experiments}

In this section, we present experimental results. Our code will be open-sourced.

\paragraph{Illustration of the Theorems.} We first illustrate the theoretical results of Theorems \ref{thm:linear_linear}, \ref{thm:stationary}, and \ref{thm:periodic}. For the three instances in \mic{Assumption} \ref{def:instances}, we compute $u^{\star}_t$ as defined in Proposition \ref{prop:discrete_causal}. We then examine the evolution of $\|u^{\star}_t - x_{t+1}\|^2$ with $t$. The corresponding curves are shown in Figure \ref{fig:illustration_theorems} (left and center), illustrating the convergence of $u^{\star}_t - x_{t+1}$ to zero for the considered instances. As predicted by our theory, the convergence is exponential for instances (1) and (3).
\vspace{-.5em}
\paragraph{More Complex Iterations.} We also explore the potential convergence of $u^{\star}_t - x_{t+1}$ to zero for a new instance (4), described as follows. 
 \mic{The key idea is to apply ``non-linear rotations'' to 2-dimensional chunks of the input. Each of these rotations is parameterized by an angle $\theta$ and a scalar $q$, and defined as $z \mapsto R_{\theta} z^q$, where $R_{\theta}$ is a 2 dimensional rotation of angle $\theta$. We now formalize this idea.} 
We consider sequences $\{x_t\}$ in $\mathbb{R}^{2p}$ generated according to the process in \mic{Assumption} \ref{def:seq}, where the mapping $f: \mathbb{R}^{2p} \to \mathbb{R}^{2p}$ is defined as follows.
We first convert the real vector $x_t$ into a complex vector $z_t \in \mathbb{C}^p$, where $z_t^{(j)} = x_t^{(2j - 1)} + i \, x_t^{(2j)}$, for $j = 1, 2, \dots, p$.
We then apply a unitary matrix $U \in \mathbb{C}^{p \times p}$ to $z_t$, defining $z_t' = U z_t$.
Next, we modulate the magnitude and phase of $z_t'$ with a scalar $q \in \mathbb{R}$ and a bias vector $\theta \in \mathbb{R}^p$ as follows: 
$$z_t'' = \exp(i \theta) \odot \left( \left| z_t' \right| \odot \exp\left( i q \arg\left( z_t' \right) \right) \right),$$ where $\odot$ denotes element-wise multiplication, $| z_t' |$ is the element-wise magnitude, and $\arg(z_t')$ is the element-wise phase of $z_t'$.
We then compute $z_{t+1} = U^{\star} z_t''$, where $U^{\star}$ is the Hermitian transpose of $U$. Finally, we convert $z_{t+1}$ back to a real vector $x_{t+1} \in \mathbb{R}^{2p}$.
The sequence starts from an initial vector $x_1 \in \mathbb{R}^{2p}$, normalized to unit length. The unitary matrix $U$ and bias vector $\theta$ are randomly initialized. The parameter $q$ controls the non-linearity applied to the phase of the transformed vector.
In Figure \ref{fig:illustration_theorems} (right), we plot $\|u^{\star}_t - x_{t+1}\|^2$ against $t$ for $d = 4$, $q = 2$, and $k = k_{\text{exp}}$. We observe convergence to zero, suggesting that our causal kernel descent method may generalize to more complex settings.

\paragraph{Training $\Mm^n$.}
\begin{wrapfigure}{r}{0.6\textwidth}
\label{fig:trained}
\vspace{-1.5em}
\includegraphics[width=0.6\textwidth]{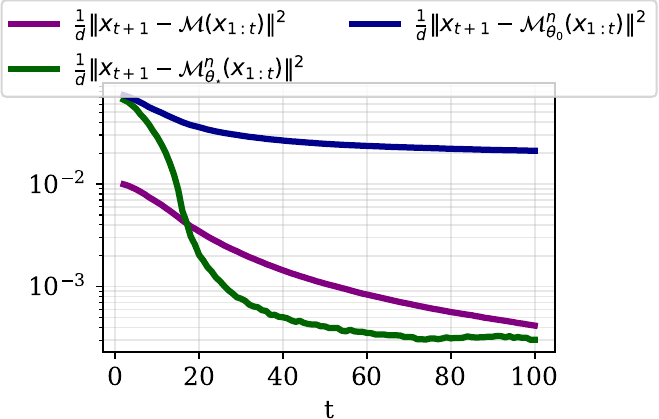}
\vspace{-2.em}
\caption{\small{Errors $\frac1d\|\Gg(x_{1:t})- x_{t+1}\|^2$} against $t$ for $\Gg \in \{\Mm, \Mm^n_{\theta_0}, \Mm^n_{\theta_{\star}} \}$. Results are averaged over the whole test set.}
\vspace{-1.2em}
\end{wrapfigure}
In Theorem \ref{thm:main}, the number of layers, $n$, is taken to infinity to obtain the estimate $u^\star_t$ of $x_{t+1}$. 
However, we experimentally demonstrate that fine-tuning the model $\Mm^n$ in \eqref{eq:discrete} leads to a trained model that competes with the infinitely deep model $\Mm$. 
For this, we take $\Nn = \text{softmax}$ to build $\Mm^n$, and fine-tune the corresponding weights. Note that, for simplicity, we consider square matrices for the parameters, by completing the parameters with zeros. To emphasize the parameter dependency, we denote by $\Mm^n_{\theta_0}$ the corresponding initialization (which therefore satisfies Theorem \ref{thm:main}).
We take $d=15$, $n=6$, and consider instance $(2)$ with randomly generated $\Omega$'s and $x_1$'s, for a dataset with $2^{12}$ elements, that we split into train, validation, and test sets with respective sizes of $60 \%$, $20 \%$, and $20 \%$ of the original dataset. We train the model using Adam \citep{kingma2014adam} on the Mean Squared Error (MSE) loss for next-token prediction on sequences of length $T=100$, i.e., we minimize:
$$
\ell(\theta) \eqdef \frac1d \sum_{t=1}^{T-1} \|\Mm_{\theta}^n(x_{1:t}) - x_{t+1}\|^2,
$$
over the parameters $\theta$ of $\Mm^n$, starting from the initialization $\theta_0$.
We train for 5000 epochs with early stopping. We denote by $\Mm^n_{\theta_{\star}}$ the corresponding model. 
We then examine how the error $\frac1d\|\Mm_{\theta_{\star}}^n(x_{1:t}) - x_{t+1}\|^2$ behaves with $t$. We find that not only does $\Mm_{\theta_{\star}}^n$ significantly outperform $\Mm_{\theta_0}^n$, but it also outperforms the infinite depth model $\Mm$ when $t \gtrsim 20$, as shown in Figure \ref{fig:trained}. 
\section*{Conclusion}
In this paper, we took a step towards understanding the universality of Transformers for next-token prediction by considering sequences of the form $ x_{t+1} = f(x_t) $ for some hidden variable $ f $. We demonstrated in Theorem \ref{thm:main} that an explicitly constructed Transformer can accurately predict the next token $ x_{t+1} $ as $ t \to +\infty $, in specific cases where $ f $ is linear or the sequence $ (x_t)_{t \geq 1} $ is periodic. Our construction corresponds to the Transformer implementing causal kernel descent methods, which provably provide consistent estimates of the next token $ x_{t+1} $ under the specific cases considered in this paper. Experimental results validated our theoretical findings and indicated that these methods can be extended to more general mappings $ f $, paving the way for future investigations to generalize our theoretical results. 

\mic{One current limitation of our approach is the deterministic aspect of the sequences considered in the paper. A possible extension of our work would be considering noisy dynamics of the form $x_{t+1} = f(x_t) + \varepsilon_t$ where the $\varepsilon_t$'s are i.i.d random variables. Another open problem concerns the ability of other sequential architectures such as RNNs or state-space models to learn in-context such autoregressive processes. These questions are left for future research.}

\section*{Acknowledgments}
The work of M. Sander and G. Peyré was supported by the
French government under the management of Agence Nationale de la Recherche as part of the “Investissements
d’avenir” program, reference ANR-19-P3IA-0001 (PRAIRIE 3IA Institute). 
The work of G. Peyré is supported by the European Research Council (ERC project WOLF).
M. Sander thanks Francisco Andrade and Pierre Marion for fruitful discussions.

\bibliography{example_paper}
\bibliographystyle{iclr2025_conference}
\newpage

\appendix
\section{Proofs}
\subsection{Proof of Proposition \ref{prop:construction} }\label{proof:construction}

\begin{proof}
Let $n > 0$ be an integer.

We define positional encodings $ p_t \in \mathbb{R} $ as
  $$
  p_t = (-1)^t n t,
  $$
  
We concatenate the input embeddings and positional encodings and define
  $$
  x^p_t = (x_t, p_t) \in \mathbb{R}^{d+1}.
  $$

We define the weight matrices $ W_Q^{(h)}, W_K^{(h)}, W_V^{(h)} $ for each head $ h = 1, 2 $ as follows:
\begin{itemize}
    \item $W_Q^{(1)} = [\mathbf{0}_{1 \times d}, \ 1], \quad W_K^{(1)} = [\mathbf{0}_{1 \times d}, \ -1],  \quad W_V^{(1)} = \begin{pmatrix}
  I_d & \mathbf{0}_{d \times 1} \\
  \mathbf{0}_{2d \times d} & \mathbf{0}_{2d \times 1} 
  \end{pmatrix}.$
  \item $W_Q^{(2)} = [\mathbf{0}_{1 \times d}, \ 1], \quad W_K^{(2)} = [\mathbf{0}_{1 \times d}, \ 1], \quad W_V^{(2)} = \begin{pmatrix}
   \mathbf{0}_{d \times d} & \mathbf{0}_{d \times 1} \\
  I_d & \mathbf{0}_{d \times 1} \\
  \mathbf{0}_{d \times d} & \mathbf{0}_{d \times 1}
  
  \end{pmatrix}.$
  
\end{itemize}

With such constructions, one has
\begin{itemize}
    \item $\langle W_Q^{(1)} x^p_{t}, W_K^{(1)} x^p_{s} \rangle = -p_s p_t$ and $W_V^{(1)}x^p_{s} = (x_s, 0, 0) $.
    \item $\langle W_Q^{(2)} x^p_{t}, W_K^{(2)} x^p_{s} \rangle = p_s p_t$ and $W_V^{(1)}x^p_{s} = (0, x_s, 0) $.
\end{itemize}

The corresponding attention scores are
    $$
    \mathcal{A}^h_{t,s} = \frac{e^{(-1)^h p_s p_t}}{\sum_{\tau=1}^{t} e^{ (-1)^h p_{\tau} p_t}} = \frac{1}{\sum_{\tau=1}^{t} e^{ (-1)^h p_{\tau} p_t - (-1)^h p_{s} p_t}}.
    $$
When $h = 1$, one has that $(-1)^h p_{s} p_t = (-1)^{t-s+1} n^2 s t $ is maximal and strictly positive when $s = t-1$. Similarly, when $h = 2$, one has that $(-1)^h p_{s} p_t = (-1)^{t-s} n^2 s t $ is maximal and strictly positive when $s = t$. Therefore, $\mathcal{A}^1_{t,s} \to \delta_{s=t-1}$ and $\mathcal{A}^2_{t,s} \to \delta_{s=t}$ as $n \to +\infty $.

    We therefore consider the forward rule defined as, for $ t \geq 1 $:
    $$
    \mathcal{F}^{n}(x_{0:T})_t = \sum_{s=1}^t \mathcal{A}^1_{t,s} \cdot (x_s, 0, 0) + \sum_{s=1}^t \mathcal{A}^2_{t,s} \cdot (0, x_s, 0).
    $$
    
    As $ n \to +\infty $, one has $ \mathcal{F}^{n}(x_{0:T})_t \to (x_{t-1}, x_t, 0) \eqdef \mathcal{F}(x_{0:T})_t $. 

We then apply a feedforward map $\Gg$ on each token $(x_{t-1}, x_t, 0)$ defined as \mic{$g(a, b, c)=(\begin{pmatrix} a \\ 1_{a=0_d} \end{pmatrix}, \begin{pmatrix} b \\ 1 \end{pmatrix}, (1-1_{a=0_d})b, c)$} to obtain the desired augmented tokens $e_t$. The feedforward map $\Gg$ can be approximated with a sigmoid based perceptron $\Gg^{n}$ using the fact that $\frac{2}{1 + e^{\|a\| n}} \to 1_{a=0_d}$ as $n \to +\infty$. 

Denoting $\mathcal{T}_0$ the Transformer composed of the attention module $\mathcal{F}$ and the feedforward module $\Gg$ conludes the proof.

\end{proof}

\subsection{Proof of Proposition \ref{prop:discrete_causal}}\label{proof:discrete_causal}
\begin{proof}
Recall that the matrix $A$ is defined as:
$$
A_{t,s}= 
\begin{cases} 
k(x_t, x_s)1_{s\leq t}, & \text{if } k = k_{\text{id}} \\
k(x_t, x_s)1_{s\leq t} \text{ or } \frac{k(x_t, x_s)}{\sum_{\tau=1}^t k(x_t, x_{\tau})}1_{s\leq t}, & \text{if } k = k_{\text{exp}}
\end{cases}.
$$

We have in matrix notations:
$$
u^{n}_{1:t} = (I_t - (I_t-\eta A)^n)A^{-1} (A-\diag(A))x_{2:t+1}.
$$

Importantly, when $A_{t,s} = k(x_t, x_s)$, for all $0 < \eta < \frac2{k(x_1,x_1)}$, one has $(I_t-\eta A)^n \to 0_{t\times t}$ exponentially fast. In this case, we therefore have $\eta^{\star} = \frac2{k(x_1,x_1)}$.
Even more, when $\eta = \frac1{k(x_1,x_1)}$, the matrix $(I_t-\eta A)$ is nilpotent, and $(I_t-\eta A)^n = 0_{t\times t}$ for $n \geq t$.

When $A_{t,s} = \frac{k(x_t,x_s)}{\sum_{\tau=1}^tk(x_t,x_{\tau})}$, for all $0 < \eta < 2$, one has $(I_t-\eta A)^n \to 0_{t\times t}$ exponentially fast. In this case, we therefore have $\eta^{\star} = 2$.

In both cases, when $0 < \eta < \eta^{\star}$, we have $\lim_{n\to+\infty}u^n_{1:t} = u_{1:t}^\star \eqdef A^{-1} (A-\diag(A))x_{2:t+1}$.
\end{proof}

\subsection{Proof of Proposition \ref{prop:mu}}\label{proof:mu}

\begin{proof}
    When $0< \eta < \eta^{\star}$, at convergence, we have:
$$
u^{\star}_{1:t} = x_{{2:t+1}} - A^{-1} \diag (A) x_{{2:t+1}}.
$$
Writing $z \eqdef A^{-1} \diag (A) x_{{2:t+1}}$ and $\mu_s \eqdef \frac{z_s}{k(x_1,x_1)}$, we obtain the recursions on $\mu$:
\begin{equation}\label{eq:dual2}
   \sum_{s=1}^t \mu_s k(x_s, x_t) = \frac{k(x_t, x_t)}{k(x_1,x_1)} x_{t+1} = W \phi(x_t),   \quad \forall t \geq 1,
\end{equation}
where the second equality comes from the fact that $\|x_t\| = \|x_1\|$. 

We have $u^{\star}_t = x_{t+1} - k(x_1, x_1)\mu_t$.
\end{proof}

\subsection{Proof of Proposition \ref{prop:diff}}\label{proof:diff}

\begin{proof}
    We have
    $$
    W_{t+1}=W_t + \mu_{t+1} \phi(x_{t+1})^*. 
    $$
    Right multiplying by  $\phi(x_{t+1})$ gives 
    $$
    \mu_{t+1}k(x_{t+1}, x_{t+1}) = x_{t+2} - W_t\phi(x_{t+1})
    $$
    Therefore, 
    $$
    W_{t+1} = W_t(I-\frac{\phi(x_{t+1})\phi(x_{t+1})^*}{k(x_{t+1}, x_{t+1})}) + \frac{x_{t+2}}{k(x_{t+1}, x_{t+1})}\phi(x_{t+1})^*.
    $$
    Since $$x_{t+2}\phi(x_{t+1})^* = W \phi(x_{t+1})\phi(x_{t+1})^*,$$
    we obtain 
    $$
    W_{t+1} = (W_t - W)(I-\frac{\phi(x_{t+1})\phi(x_{t+1})^*}{k(x_{t+1}, x_{t+1})}) + W 
    $$
    
Since $ W_1 = W \nu_1 \nu_1^* $, it follows that $ W_t - W = - W P_1 P_2 \cdots P_t $, where $ P_s =  \left(I - \nu_s \nu_s^* \right) $ is the orthogonal projection onto the subspace orthogonal to $ \nu_s $.
    \end{proof}

\subsection{Proof of Theorem \ref{thm:linear_linear} }\label{proof:linear_linear}
    
    \begin{proof}
 We first prove the second statement of the theorem, that is that $W_t - W \to 0$ exponentially fast for almost all $x_1$ and $W$.      
 
 Let's first fix $W$ and $x_1$. In what follows we denote $x \eqdef x_1$ to ease the notations.
 
    Let $\Delta_t = W_t - W$. We have 
    $$
    \Delta_{t+1} = \Delta_t W^t(I-xx^\top) W^{-t}.
    $$
Unrolling, we obtain 
$$
\Delta_{t+1} = \Delta_1 (W (I-xx^\top))^t W^{-t} = - (W (I-xx^\top))^{t+1} W^{-t},
$$
because $\Delta_1 = - W (I-xx^\top)$.

For fixed $W$ and $x$, let $y$ be an eigenvector of norm $1$ of $W (I-xx^\top)$. There exists $\lambda$ such that 
$$
W (I-xx^\top) y= \lambda y.$$
If $|\lambda|=1$, then because $W$ preserves the norm, it follows that $$\|(I-xx^\top) y\|^2 = \|y\|^2 = 1.$$
Developing, we get $$\|x\langle x , y \rangle\|^2 = 0.$$
But since $$\|x\langle x , y \rangle\|^2 =  \langle x , y \rangle^2, $$ we must have $\langle x , y \rangle = 0$, and therefore $W y = \lambda y$, so that $\lambda$ is also an eigenvalue of $W$.

Therefore, to show that for almost all $x$ and $W$, $\rho(W(I-xx^\top)) <1$, it suffices to show that for almost all $y \in V(W)$:
$$ \langle x , y \rangle  \neq 0,$$
where $V(W)$ denotes the eigen vectors of unit norms of $W$. 

Let $DO(d)$ be the subset of $O(d)$ of orthogonal matrices with distinct eigenvalues and let $W \in DO(d)$.

Let $\lambda_1, \cdots, \lambda_d$ be the eigenvalues of $W$ and $b_1, \cdots, b_d$ be $d$ corresponding eigenvectors. Let $y \in V(W)$, and $\lambda$ be the corresponding eigenvalue. 

Writing $y = \sum_{i=1}^d y_i b_i$, one has on the one hand that 
$$Wy = \lambda y $$ and on the other hand $$Wy= \sum_{i=1}^d y_i \lambda_i b_i.$$ 
Identifying the coefficients gives $\lambda = \lambda_i$ whenever $y_i \neq 0$. Since the $\lambda_i$'s are all distinct, necessarily exactly one $y_i$ is non-zero, and $y \in \CC b_i$.

Therefore, if $ \langle x , y \rangle  =0$, then $x \in \cup_{i=1}^d (\CC b_i)^{\perp}$. However, for almost all $x \in S^{d-1}$,  $x \notin \cup_{i=1}^d (\CC b_i)^{\perp}$. Therefore, for almost all $W \in DO(d)$ and $x \in S^{d-1}$, 
$$ \langle x , y \rangle  \neq 0.$$

Since for almost all $W \in O(d)$, $W \in DO(d)$, we conclude that $\rho(W(I - xx^\top)) < 1$ for almost all $x \in S^{d-1}$ and $W \in O(d)$. As a consequence, for almost all $x \in S^{d-1}$ and $W \in O(d)$, $W_t - W \to 0$ as $t \to +\infty$.
It already implies that, for almost all $x$ and $W$, $\mu_t x^\top_t \to 0$. Therefore, $\text{Tr}(\mu_t x^\top_t x_t \mu_t^\top) \to 0$. 
But since $x_t^\top x_t = 1$ and  $\text{Tr}(\mu_t \mu_t^\top) = \|\mu_t\|$, it gives $\mu_t \to 0$, and therefore $u_t^\star - x_{t+1} \to 0$ for almost all $x_1$ and $W$.

However, to fully prove the first statement of the theorem, that is that, \textbf{for all} $x_1$ and $W$, $u_t^\star - x_{t+1} \to 0$, we follow the exact same demonstration as for Theorem \ref{thm:stationary} (corresponding to the next proof in Appendix \ref{proof:stationary}), simply by replacing exp by id in \eqref{eq:g}.
\end{proof}

\subsection{Proof of Theorem \ref{thm:stationary}}\label{proof:stationary}
\begin{proof}
As mentioned in the proof sketch, we consider the subset of $ \Hh $ of linear combinations of the $\nu_s$:
$$
x \mapsto \sum_{s=1}^{\tau} a_s \nu_s, \text{ for }   a_s \in \RR^d\text{, } \tau \geq 1.
$$

This subset is a pre-Hilbert space under the inner product inherited from $ \Hh $. By completing this space with respect to the induced norm from $\Hh$, we obtain a new Hilbert space $ \Hh' $

We first show that $ P_t P_{t-1} \cdots P_1 \to 0 $ strongly in $\Hh'$, using  a characterization from \citet{kwapien2001kaczmarz}.

A sequence $ (\nu_s)_{s \geq 1}$ of unit vector for which $ P_t P_{t-1} \cdots P_1 \to 0 $ strongly is referred to as \textit{effective}.

The specificities of our case are twofold: first, $ \nu_s  \in \mathcal{H'} $, which is potentially of infinite dimension, and second, the vectors $ \nu_s $ follow an autoregressive relation.
In our case, we can still show that the sequence $(\nu_s)_{s \geq 1}$ is effective in $\Hh'$. 

Note that because $\Omega\in O(d)$, one has for any positive integers $t, s, r$ that $\langle \nu_{s + r}, \nu_{t + r} \rangle = \langle \nu_{s}, \nu_{t} \rangle$. Such sequence is called \textit{stationary}.

Bochner's theorem states that there is a measure $\sigma$ on the unit circle $S^1$--called \textit{spectral measure}--such that, for all $t \geq 1$,
$$
 a_t \eqdef \langle \nu_{t+1}, \nu_1 \rangle = \int_{S^1} z^t \, d\sigma(z).
$$
We are going to use the following characterization from \citet{kwapien2001kaczmarz, RainisHaller2005}.
\paragraph{Theorem}(Effectiveness of stationary sequences \citep{kwapien2001kaczmarz}).
\textit{A stationary sequence of unit vectors which is linearly dense in a Hilbert space is effective if and only if its spectral measure either coincides with the normalized Lebesgue measure or is singular with respect to the Lebesgue measure.}

Now, let $\Omega$ be an orthogonal matrix.
One has 
$$a_t =\langle \nu_{t+1}, \nu_1 \rangle = \frac{1}{k(x_1, x_1)} k(x_{t+1}, x_1) =  \frac{1}{k(x_1, x_1)} \exp{(\langle x_1, \Omega^t x_1 \rangle)}.
$$
Since $\Omega$ is an orthogonal matrix, it can be diagonalized via rotations with angles $\theta_1, \theta_2, \ldots, \theta_p$.

We therefore can write $\langle x_1, \Omega^t x_1 \rangle$ as

$$
\langle x_1, \Omega^t x_1 \rangle = \sum_{i=1}^p (x_{1(2i-1)}^2 + x_{1(2i)}^2)  \cos(\theta_i t) + \sum_{i=2p+1}^m x_{1i}^2 - \sum_{i=m+1}^d x_{1i}^2, 
$$
for some $p$ and $m$.
Therefore, one can write
$$
a_t = g(t(\theta_1, \cdots, \theta_p)),
$$
where $g$ is defined as 
\begin{equation}\label{eq:g}
g(y_1, \cdots, y_p) \eqdef \exp{\left(-1 +  \sum_{i=1}^p (x_{1(2i-1)}^2 + x_{1(2i)}^2)  \cos(y_i) + \sum_{i=p+1}^m x_{1i}^2 - \sum_{i=m+1}^d x_{1i}^2\right)}.
\end{equation}

The function $g$ is periodic in each variable $y_i$ with period 
$2\pi$. Since $g$ is also $C^1$, this allows us to represent $g$ as a Fourier series:
$$
g(y) = \sum_{k \in \ZZ^p} c_k e^{i\langle y, k \rangle}.
$$
Therefore, 
$$a_t = \sum_{k \in \ZZ^p} c_k e^{i t \langle \theta, k \rangle}.$$ 

Therefore, the spectral measure satisfies 
$$
\sigma = \sum_{k \in \ZZ^p} c_k \delta_{\langle \theta, k \rangle}.
$$ 
Note that this necessarily implies that $c_k \geq 0$ by Bochner's theorem. We are now going to prove that $\sigma$ is singular with respect to the Lebesgue measure.

We recall that two positive measures $\alpha$ and $\beta$ defined on a measurable space $C$ are singular if there exist two disjoint measurable sets $A, B$ such that $A \cup B = C$, such that $\alpha$ is zero on all measurable subsets of $B$, while $\nu$ is zero on all measurable subsets of $A$. Here, $\alpha = \sigma$, and $\beta$ is the Lebesgue measure on $C = S^1$. 

We have $$A = \cup_{k\in \ZZ^p}\{\langle \theta , k \rangle \}.$$

$A$ is measurable and the Lebesgue measure $\beta$ is zero on $A$, while by definition, $\sigma$ is zero on all measurable subsets of $B \eqdef S-A $. 

Therefore, $\sigma$ is singular with respect to the Lebesgue measure, which, using Theorem 2 from \cite{kwapien2001kaczmarz} in the Hilbert space $\Hh'$ shows that the sequence $(\nu_s)_{s \geq 1}$ is effective in $\Hh'$. 

Now, we have, by taking the adjoint $^*$ of $W_t - W$:
$$
W^*_t - W^* =- P_t \cdots P_1 W^*.
$$

We recall that we identify $W$ with $W_{\Hh'}$ so that $W : \Hh' \to \RR^d$ and $W^* : \RR^d \to \Hh'$.
One has for all $x \in \RR^d$
$$
(W^*_t - W^*)x = \sum_{s=1}^t \phi(x_s) \langle z_s , x \rangle - W^*x = - P_t \cdots P_1 W^*x \to 0
$$
because $W^* x \in \Hh'$. 

Therefore, for all $x \in \RR^d$, $\phi(x_t) \langle z_t , x \rangle \to 0$ in $\Hh'$. Because $\| \phi(x_t)\|_{\Hh'} = k(x_1,x_1) > 0$, we necessarily have $\langle z_t , x \rangle \to 0$ for all $x \in \RR^d$. This is equivalent to $z_t \to 0$, and therefore $u_t^\star - x_{t+1} \to 0$.

\end{proof}

\subsection{Proof of Theorem \ref{thm:periodic}}\label{proof:periodic}
\begin{proof}
 Let $t_p$ denotes the period of $(x_t)_{t\geq1}$. 
We define 
$$\Pi \eqdef  P_1 \cdots P_{t_p}.$$
One has $\sup_{\nu \in \Hh'\text{, } \|\nu\| = 1} \|\Pi \nu\| < 1$ (see \citet{RainisHaller2005}, page 2).
Therefore, $\Pi^m \to 0$ as $m \to +\infty$.

 As such, for any $x$ such that $\phi(x) \in \Hh'$, $W_t \phi(x) \to W \phi(x)$ exponentially fast. 
 
 In particular, for $x = x_1$, we get that 
 $$
 \sum_s \mu_s k(x_s, x_1) 
 $$
 converges exponentially fast. Since $k(x_s, x_1) > \exp({-1})$, it implies that $\mu_s \to 0$ exponentially fast.
\end{proof}

\subsection{Proof of Proposition \ref{prop:discrete}}\label{proof:discrete}

\begin{proof}

\textbf{Building positional encodings and parameters.}

    Similarly to the previous proof, we consider positional encodings $p^1_t$ and $ p^2_t \in \mathbb{R} $ as
  $$
  p^1_t =\delta_{s=1}, \quad p^2_t = 1.
  $$
 
We concatenate the input embeddings and positional encodings and define
  $$
  e_{t,p} = (x_{t-1} , p^1_t , x_{t}, p^2_t , x_t, u_t) =  (x_{t-1}, 0 ,  x_{t},1, x_t, u_t) \in \mathbb{R}^{4d+2}
  $$
if $t >1$ and 
$$
e_{1,p} = ( 0_d , p^1_1,  x_1, p^2_1, 0_d, u_1) = ( 0_d , 1 , x_1, 1 , 0_d, u_1).
$$ 

We define the weight matrices $ W_Q^{(h)}, W_K^{(h)}, W_V^{(h)} $ for each head $ h = 1, 2 $ as follows:
\begin{itemize}[left=0pt]
    \item $W_Q^{(1)} =  W_K^{(1)} = [\mathbf{0}_{d \times d+1},I_d, \mathbf{0}_{d \times 2d+1}],$ and 
    
    $W_V^{(1)} = - \eta \begin{pmatrix}
 \mathbf{0}_{d+1 \times d+1} & \mathbf{0}_{d+1\times d+1} & \mathbf{0}_{d +1 \times d} & \mathbf{0}_{d +1 \times d} \\
 \mathbf{0}_{d+1 \times d+1} & \mathbf{0}_{d+1\times d+1} & \mathbf{0}_{d +1 \times d} & \mathbf{0}_{d +1 \times d} \\
  \mathbf{0}_{d \times d+1} & \mathbf{0}_{d\times d+1} & \mathbf{0}_{d \times d} & \mathbf{0}_{d \times d} \\
\mathbf{0}_{d \times d+1} & \mathbf{0}_{d\times d+1} & \mathbf{0}_{d \times d} & I_d \end{pmatrix}$.
   
  \item $W_Q^{(2)} = [\mathbf{0}_{d+1 \times d+1}, I_{d+1},\mathbf{0}_{d+1 \times 2d}],$ $W_K^{(2)} = [I_{d+1}, \mathbf{0}_{d+1 \times d+1},\mathbf{0}_{d+1 \times 2d}],$ $W_V^{(2)} =  \eta  \begin{pmatrix}
\mathbf{0}_{d+1 \times d+1} & \mathbf{0}_{d+1 \times d+1} & \mathbf{0}_{d+1 \times d}& \mathbf{0}_{d+1 \times d} \\
\mathbf{0}_{d+1 \times d+1} & \mathbf{0}_{d+1 \times d+1} & \mathbf{0}_{d+1 \times d}& \mathbf{0}_{d+1 \times d} \\
  \mathbf{0}_{d \times d+1} & \mathbf{0}_{d\times d+1} & \mathbf{0}_{d \times d}& \mathbf{0}_{d \times d} \\
  \mathbf{0}_{d\times d+1} & \mathbf{0}_{d\times d+1} & I_d & \mathbf{0}_{d \times d}
    \end{pmatrix}$.
  
\end{itemize}

With such constructions, one has
\begin{itemize}[left=0pt]
    \item $\langle W_Q^{(1)} e_{t,p}, W_K^{(1)} e_{s,p} \rangle = \langle x_t, x_s \rangle$ and $W_V^{(1)}e_{s,p} = - \eta (0_{d+1}, 0_{d+1}, 0_d, u_s)$.
    \item $\langle W_Q^{(2)} e_{t,p}, W_K^{(2)} e_{s,p} \rangle = \langle x_t, x_{s-1} \rangle + p^1_s p^2_t = \langle x_t, x_{s-1} \rangle + \delta_{s=1}$ and $W_V^{(2)}e_{s,p} =  \eta  (0_{d+1}, 0_{d+1}, 0_{d}, x_s) $ if $s > 1$ and $W_V^{(2)}e_{1,p} = (0_{d+1}, 0_{d+1}, 0_{d}, 0_d)$.
\end{itemize}

\textbf{Implementing \eqref{eq:discrete_causal} when $A_{t,s} = k(x_t,x_s)$.}

Let $\Tt$ be a one-layer, two-head Transformer with the parameters above (with $\Nn=\text{id}$ of $\Nn = \text{exp}$).
Its forward rule is defined as, starting from $e^0_{t,p} = e_{t,p}$:
$$
e^{k+1}_{t,p} = e^k_{t,p}-\eta \sum_{s=1}^t \mathcal{A}^1_{s,t} (0_{d+1}, 0_{d+1}, 0_d, u^k_s)  +\eta \sum_{s=1}^t\mathcal{A}^2_{s,t} (0_{d+1}, 0_{d+1}, 0_d, x_s). 
$$
Therefore, the $(3d+2)$ first coordinates of $e^k_{t,p}$ are not modified and the $d$ last coordinates $u^k_t$ are updated as
\begin{equation}\label{eq:update1}
u^{k+1}_t = u^{k}_t  - \eta \sum_{s=1}^tk(x_t, x_s)u^k_s + \eta \sum_{s=2}^tk(x_t, x_{s-1})x_s
\end{equation}
The second term in \eqref{eq:update1} reads 
$$
\sum_{s=2}^tk (x_t, x_{s-1})x_s = \sum_{s=1}^{t-1}k(x_t, x_{s})x_{s+1}.
$$
Therefore, \eqref{eq:update1} reads
$$
u^{k+1}_t = u^{k}_t - \eta \sum_{s=1}^{t}k(x_t, x_{s}) (u^k_s - 1_{s < t }x_{s+1}),
$$
which corresponds exactly to \eqref{eq:discrete_causal}.

\textbf{Implementing \eqref{eq:discrete_causal} when $A_{t,s} =\frac{k(x_t,x_s)}{\sum_{\tau=1}^t k(x_t, x_{\tau})} $.}

Let $\Tt$ be a $\Nn=\text{softmax}$-based, one-layer, two-head Transformer with the parameters above. Its forward rule is defined as 
$$
e^{k+1}_{t,p}= e^k_{t,p}-\eta \sum_{s=1}^t \mathcal{A}^1_{s,t} (0_{d+1}, 0_{d+1}, 0_d, u^k_s)  + \eta \sum_{s=1}^t\mathcal{A}^2_{s,t} (0_{d+1}, 0_{d+1}, 0_d, x_s). 
$$
Therefore, the $(3d+2)$ first coordinates of $e_{t,p}$ are not modified and the $d$ last coordinates $u_t$ are updated as
\begin{equation}\label{eq:update}
u^{k+1}_t = u^k_t - \eta \sum_{s=1}^t\frac{e^{\langle x_t, x_s \rangle}}{\sum_{\tau=1}^{t} e^{ {\langle x_t, x_\tau \rangle}}}u^k_s  + \eta \sum_{s=2}^t\frac{e^{\langle x_t, x_{s-1} \rangle + \delta_{s=1}}}{\sum_{\tau=1}^{t} e^{ \langle x_t, x_{\tau-1} \rangle + \delta_{\tau=1}}} x_s.
\end{equation}
Because $\|x_t\|=1$ for all $t \geq 1$ and $x_0 = 0$, one has 
$$
\sum_{\tau=1}^{t} e^{ \langle x_t, x_{\tau-1} \rangle + \delta_{\tau=1}}= \sum_{\tau=1}^{t} e^{ \langle x_t, x_{\tau} \rangle }
$$
so that the second term in \eqref{eq:update} reads 
$$
\sum_{s=2}^t\frac{e^{\langle x_t, x_{s-1} \rangle + \delta_{s=1}}}{\sum_{\tau=1}^{t} e^{ \langle x_t, x_{\tau-1} \rangle + \delta_{\tau=1}}} x_s = \sum_{s=2}^t\frac{e^{\langle x_t, x_{s-1} \rangle }}{\sum_{\tau=1}^{t} e^{ \langle x_t, x_{\tau} \rangle }} x_s = \sum_{s=1}^{t-1}\frac{e^{\langle x_t, x_{s} \rangle }}{\sum_{\tau=1}^{t} e^{ \langle x_t, x_{\tau} \rangle }} x_{s+1}.
$$
Therefore, \eqref{eq:update} reads
$$
u^{k+1}_t = u^{k}_t - \eta \sum_{s=1}^{t} \frac{e^{\langle x_t, x_{s} \rangle }}{\sum_{\tau=1}^{t} e^{ \langle x_t, x_{\tau} \rangle }} (u^k_s - 1_{s < t }x_{s+1}),
$$
which corresponds exactly to \eqref{eq:discrete_causal}.

\end{proof}

\subsection{Proof of Theorem \ref{thm:main}}\label{proof:main}
\begin{proof}
    We first use Proposition \ref{prop:discrete}. For any $0 < \eta < \eta^\star$, where $\eta^\star$ of defined in Proposition \ref{prop:discrete_causal} and for each configuration of \eqref{eq:discrete_causal}, there exists an attention-only, one-layer, two-head causal Transformer $\Tt$ such that, for any autoregressive sequence $(x_{t})_{t\geq 1} $ generated according to \mic{Assumption} \ref{def:seq}, defining $$e^{k}_t \eqdef (x_{t-1} , 0, x_t, 1, x_t, u^k_t) $$ for $t>1$ and $$e^k_1 \eqdef (0_{d}, 1, x_t, 1, 0_d, u^k_t),$$ 
    
    we have that, for any $n$, $e_{1:t}^{1:n}$ solves \eqref{eq:discrete} if and only if $u_{1:t}^{1:n}$ solves \eqref{eq:discrete_causal}. 

    Thanks to Proposition \ref{prop:discrete_causal}, we know that $u^{n}_t \to_{n\to+\infty} u^\star_t$, and define $\Mm(x_{1:t}) \eqdef u^\star_t$. Then:
    \begin{itemize}[left=0pt]
        \item When $A_{t,s} = \langle x_t, x_s \rangle$, taking $\Nn = \text{id}$ in the construction of $\Tt$ in the proof of Proposition \ref{prop:discrete}, and under instance $(1)$, we have $\Mm(x_{1:t}) -  x_{t+1} \to 0$, with exponential speed for almost all $x_1$ and $W$, thanks to Theorem \ref{thm:linear_linear}.
        \item When $A_{t,s} = e^{ \langle x_t, x_s \rangle}$ (resp. $A_{t,s} = \frac{e^{ \langle x_t, x_s \rangle}}{\sum_{\tau=1}^te^{ \langle x_t, x_{\tau} \rangle}}$), taking $\Nn = \text{exp}$ (resp. $\Nn = \text{softmax}$) in the construction of $\Tt$ in the proof of Proposition \ref{prop:discrete}, and under instance $(2)$, we have $\Mm(x_{1:t}) -  x_{t+1} \to 0$  thanks to Theorem \ref{thm:stationary}.
        \item When $A_{t,s} = e^{ \langle x_t, x_s \rangle}$ (resp. $A_{t,s} = \frac{e^{ \langle x_t, x_s \rangle}}{\sum_{\tau=1}^te^{ \langle x_t, x_{\tau} \rangle}}$), taking $\Nn = \text{exp}$ (resp. $\Nn = \text{softmax}$) in the construction of $\Tt$ in the proof of Proposition \ref{prop:discrete}, and under instance $(3)$, we have $\Mm(x_{1:t}) -  x_{t+1} \to 0$, with exponential speed, thanks to Theorem \ref{thm:periodic}.
    \end{itemize}

The last statement of Theorem \ref{thm:main} follows from Proposition \ref{prop:discrete_causal}. When $\Nn=\text{id}$ or $\Nn=\text{exp}$, by choosing $\eta = \frac{1}{k(x_1,x_1)}$, we have that $\Mm^n(x_{1:t}) = u^n_t = u^\star_t = \Mm(x_{1:t})$ whenever $n \geq t$. 

When $\Nn = \text{softmax}$ however, choosing for instance $\eta = 1$ in the proof of Proposition \ref{proof:discrete_causal}, we get $u_{1:t}^n -u_{1:t}^\star = (I_t - A)^n u_{1:t}^\star$.
Since $\rho(I_t - A) \leq 1 - \frac1t$, it is reasonable to conjecture that $\| u_{t}^n -u_{t}^\star \|$ is of order $(1 - \frac1t)^n$. In this case, it is sufficient to impose $\frac{t}{n} \to 0$ as $t,n \to +\infty$ to have $\eps_1(n,t) \eqdef \| u_{t}^n -u_{t}^\star \| \to 0$ as $n,t \to +\infty$.
\end{proof}

\section{Additional results}\label{app:additional}

\paragraph{Dual interpretation.}
We provide a dual interpretation for $\mu$ defined in Proposition \ref{prop:mu}. We conveniently write \eqref{eq:OLS} as 
$   
 \min_W \|\Phi(W) -  x_{{2:T+1}}\|^2 
$
where for $\mu $ in $\mathbb{R}^{d \times T-1}$, 
$ \Phi(W) \eqdef ( W \phi(x_t) )_t $ and $
    \Phi^\top(\mu ) =  \sum_t \mu _t \phi(\mu _t)^\top $.
Gradient flow on $E$ reads:
$$
    \dot{W} = - \nabla E(W) = -[ \Phi^\top( \Phi(W) - x_{{2:T+1}} )
    ].
$$
At optimality,  one has
    $W = \Phi^T(\mu)$
for some $\mu  \in \RR^{d \times T-1}$ (``kernel trick''),  so that we consider the energy
$    F(\mu ) = E( \Phi^T(\mu) )$.
Gradient flow on $F$ reads:
 $
    \dot{\mu } = -K [ K \mu  - x_{{2:T+1}} ]
$ where 
    $K := \Phi \Phi^\top : \mu  \mapsto \left( \sum_s k(x_s,x_t) \mu_s \right)_t$.
Since $K$ is positive,  we instead consider the equivalent flow
$$
    \dot{\mu } = -[K \mu  - x_{{2:T+1}} 
    ]. 
$$
Making this flow causal by replacing $K$ with its masked counterpart $A$
recovers to the results of the previous paragraph, since, at convergence, $\mu$ satisfies \eqref{eq:dual}.

\newpage 

\end{document}

\section{Transformers for Autoregressive In-Context Learning}%

\begin{thm}[Infinite depth autoregressive in-context learning.]
For each instance in Definition \ref{def:instances}, there exists an attention-only, one-layer, two-head causal Transformers $\Tt$ such that, for any autoregressive sequence $ (x_{1:T}) $ generated according to Definition \ref{def:seq}, $ \lim_{t\to+\infty}(\Mm(x_{1:t}) - x_{t+1})=0 $, where $\Mm$ is defined in \eqref{eq:anode}. For instance $(1)$, the convergence is almost sure and exponentially fast.  For instance $(3)$, the convergence is exponentially fast. 
\end{thm}

\begin{prop}[Finite depth autoregressive in-context learning.]
\label{prop:discrete_legacy}
For each Transformer in Theorem \ref{thm:main}, there exists a step size $ \eta^{\star} $ in \eqref{eq:discrete} such that for any autoregressive sequence $ (x_{1:T}) $ generated according to Definition \ref{def:seq}, 
for all $ 0 \leq \eta \leq \eta^{\star} $, $ P e^N_t \xrightarrow[N \to \infty]{} \Mm(x_{1:t}) $ exponentially fast.
\end{prop}

We now state a proposition demonstrating that for well-chosen step sizes $ \eta $, $ \Mm^n(x_{1:t})$ is either equal or converges exactly to $ \Mm(x_{1:t}) $, the neural ODE model output in \eqref{eq:anode}.

In this section, we demonstrate how the causal kernel descent method introduced in Section \ref{sec:causal_kernel_descent} can be implemented using Transformers. Specifically, we construct a Transformer model whose output aligns with the solution of the causal flow given by \eqref{eq:modified_flow_prediction} and \eqref{eq:modified_flow_prediction_normalized}. We begin by detailing the construction of augmented tokens, which are essential for encoding the necessary information into the Transformer. Then, we show how an infinite-depth Transformer can be interpreted as a neural ODE that implements our causal flow. Finally, we discuss practical considerations for approximating this behavior with finite-depth models.

\paragraph{Augmented tokens.}

A first crucial step in our construction is to build augmented tokens given a sequence $(x_1, \cdots, x_T)$. In \citet{vonoswald2023uncovering} and \citet{sander2024transformers}, augmented tokens of the form $(x_{t-1}, x_t, 0_d) \in \RR^{3d}$ are used. These tokens explicitly encode the relative positions of the tokens $x_t$.
This approach is akin to the in-context learning literature, where sequences are often constructed as pairs $(x_i, f(x_i))$ to provide both input and output information. However, such a construction with attention based models is non trivial. \citet{sander2024transformers} proposes a way of building such tokens with general positional encoding. We here provide a more detailed result showing the augmented tokens can be computed with a one layer Transformer with dot-product absolute positional encoding. We introduce a "beginning of sequence" token $x_0 \eqdef 0_d$ and consider the extended sequence $x_{0:T} = (x_0, x_1, \cdots, x_T)$. We have the following proposition.
\begin{prop} %
    Consider the sequence $ x_{0:T} \eqdef (0_d, x_1, \cdots, x_T) $. There exists a one-layer and $2$-heads causal Transformer $ \mathcal{T}_0 $ with  $\Nn=\text{softmax}$, such that, for any $ t > 1 $, $ \mathcal{T}_0(x_{0:T})_t = (x_{t-1}, x_t, 0_d) $, and $\mathcal{T}_0(x_{0:T})_1 =  (0_d, 0_d, 0_d)$.
\end{prop}

The full proof along with the explicit construction of the corresponding model is given in Appendix \ref{proof:construction}. From now on, we consider the new augmented tokens $\mathcal{T}_0(x_{0:T})_t$ (note that $\mathcal{T}_0(x_{0:T})_1$ acts as a ``beginning of sequence'' token). We will now show how to implement \eqref{eq:modified_flow_prediction} or  \eqref{eq:modified_flow_prediction_normalized} with a neural ODE Transformer model.

\subsection{Kernel neural flows}

With the augmented tokens in place, we now demonstrate how an infinite-depth Transformer can be viewed as solving a neural ordinary differential equation (ODE) that corresponds to our causal flow equations.

An infinite stack of identical causal Transformer layers can be interpreted as a continuous-time neural ODE \citep{chen2018neural}. Specifically, considering the sequence of augmented tokens $(e^0_t)_{t=1}^T$ with $e^0_t \eqdef \mathcal{T}_0(x_{0:T})_t$, a causal Transformer dynamics can be written as
\begin{equation}%
\forall t \geq 1, \quad \frac{de}{d\tau}_t = \Tt(e_{1:t}) \quad \text{with} \quad e_t(0) = e_0^t.
\end{equation}
The output of the model is the solution at infinite time of \eqref{eq:anode}.
We denote $e_t(.) = (x_{t-1}, x_t, u_t(.))$ for $t>1$ and  $e_1 = (0_d, 0_d, u_1(.))$, with $u_t(.) = 0_d$.  
Our objective is to design $\Tt$ such that the neural ODE \eqref{eq:anode} evolves the tokens $e_t$ such that each component $u_t$ of $e_t$ solves the causal flow equations \eqref{eq:modified_flow_prediction} or \eqref{eq:modified_flow_prediction_normalized}.
 We have the following theorem. 
\begin{thm}%
    There exist three one-layer, two-head causal Transformers: $\Tt_{\text{id}}$ (with $\Nn = \text{identity}$), $\Tt_{\text{exp}}$ ($\Nn = \text{exp}$), and $\Tt_{\text{softmax}}$, such that \eqref{eq:anode} can be written on $u_t$ as \eqref{eq:modified_flow_prediction} for $\Tt= \Tt_{\text{id}}$ when $k=k_{\text{id}}$ or for $\Tt=\Tt_{\text{exp}}$ when $k=k_{\text{exp}}$  and as \eqref{eq:modified_flow_prediction_normalized} when $\Tt=\Tt_{\text{softmax}}$ and $k=k_{\text{exp}}$.
\end{thm}

The full proof along with the explicit construction of the corresponding models is given in Appendix \ref{proof:anode}.

\subsection{Approximation with finite depth Transformers}

In practice, the neural ODE in \eqref{eq:anode} is to be discretized to obtain a finite depth model. We therefore introduce a step size $\eta$ and consider the iterates, starting from $\Lambda_0 = 0$ 
$$
\Lambda_{n+1} = \Lambda_n  - \eta A \Lambda_n + \eta B.
$$
Note that this corresponds to a residual neural network where the residuals are linear and given by $- \eta A \Lambda_n + \eta B$.
The corresponding closed-form expression for the sequence $(\Lambda_n)$ is 
$$
\Lambda_n = (I -(I-\eta A)^n) A^{-1} B.
$$
The convergence and speed of convergence of $\Lambda_n$ to $\Lambda^\star = A^{-1}B$ depends solely on the matrix $N_{\eta} \eqdef (I-\eta A)$:  

\begin{itemize}
    \item Any $0 < \eta < 2$ leads to $N_{\eta}$ having spectral radius smaller than $1$ and ensures that $\Lambda_n \to_{n\to +\infty} \Lambda^\star$. This leads to exponential convergence in depth.
     \item The specific choice $\eta = 1$ leads to $N_{1}$ being nilpotent. As such, one has  $\Lambda_T = \Lambda^\star = A^{-1}B$. In this case, one needs $T$ attention layers to exactly match the output of the neural gradient flow \eqref{eq:agf_unit}.
    \item Any $\eta \geq 2$ leads to a diverging sequence.
\end{itemize}

\newpage
\newpage

\section{Draft}
\newpage 

\mic{
Any $\lambda \in Sp(W(I-xx^\top))$ satisfies
$$
|\lambda|^2 = \|(I-xx^\top)y\|^2.  
$$
Since for any $x \in S^{d-1}$, $\EE_W (\|(I-xx^\top)y\|^2) = 1 - \frac1d$, we obtain the inequality
 $$
    \EE_{W,x} \rho(W (I - x x^\top)) \leq (1 - \frac{1}{d})^{\frac{1}{2}}.
    $$
}
\begin{proof}
    One has 
    $$
    W_{T+1}=W_T + \lambda_{T+1} x^\top_{T+1}, 
    $$
    and 
    $$
    \lambda_{T+1} = x_{T+2} - W_Tx_{T+1}
    $$
    Therefore, 
    $$
    W_{T+1} = W_T(I-x_{T+1}x_{T+1}^\top) + x_{T+2}x_{T+1}^\top.
    $$
    Since $x_{T+2}x_{T+1}^\top = W x_{T+1}x_{T+1}^\top = W W^Txx^\top W^{-T}$,
    we obtain 
    $$
    W_{T+1} = (W_T - W)(I-x_{T+1}x_{T+1}^\top) + W = (W_T - W)W^T(I-xx^\top) W^{-T} + W.
    $$
    Let $\Delta_T = W_T - W$. We have 
    $$
    \Delta_{T+1} = \Delta_T W^T(I-xx^\top) W^{-T}.
    $$
Unrolling, we obtain 
$$
\Delta_{T+1} = \Delta_1 (W (I-xx^\top))^T W^{-T} = - (W (I-xx^\top))^{T+1} W^{-T}
$$
because $\Delta_1 = - W (I-xx^\top)$.
For fixed $W$ and $x$, let $y$ be an eigenvector of norm $1$ of $W (I-xx^\top)$. There exists $\lambda$ such that $W (I-xx^\top) y= \lambda y$. If $|\lambda|=1$, then because $W$ preserves the norm, it follows that $\|(I-xx^\top) y\|^2 = \|y\|^2 = 1.$ Developing, we get  $\|x\langle x , y \rangle\|^2 = 2 \langle y, x\langle x , y \rangle\rangle$. Since $\|x\langle x , y \rangle\|^2 = y^\top x x^\top x \langle x , y \rangle = y^\top x \langle x , y \rangle $, we must have $\langle x , y \rangle = 0$, and therefore $W y = \lambda y$. Therefore, to show that $\rho(W(I-xx^\top)) <1$ with probability $1$, it suffices to show that
$$
\PP_{W,x}(\exists  y \in V(W), y^\top x =0) = 0,
$$
where $V(W)$ denotes the eigen vectors of unit norms of $W$. Let $DO(d)$ be the subset of $O(d)$ of orthogonal matrices with distinct eigenvalues and let $W \in DO(d)$. Let $e_1, \cdots, e_d$ be $d$ corresponding eigenvectors. If $\exists  y \in V(W)$ and $ y^\top x =0$, then $y \in \cup_{i=1}^d (\CC e_i)^{\perp}$. But $P_x(\cup_{i=1}^d (\CC e_i)^{\perp})=0$. Therefore, for any $W \in DO(d)$, 
$$
\PP_{W,x}(\exists  y \in V(W), y^\top x =0| W) = 0.
$$
Since $DO(d)$ has full measure in $O(d)$, we conclude that $
\PP_{W,x}(\exists  y \in V(W), y^\top x =0) = 0
$.

\mic{
Any $\lambda \in Sp(W(I-xx^\top))$ satisfies
$$
|\lambda|^2 = \|(I-xx^\top)y\|^2.  
$$
Since for any $x \in S^{d-1}$, $\EE_W (\|(I-xx^\top)y\|^2) = 1 - \frac1d$, we obtain the inequality
 $$
    \EE_{W,x} \rho(W (I - x x^\top)) \leq (1 - \frac{1}{d})^{\frac{1}{2}}.
    $$
}\end{proof}
\paragraph{Autoregressive models.}

We consider Vector Autoregressive (VAR) models (e.g. \citet{hamilton2020time}, chapter 10.1) or order 1. VAR models are vector generalization of autoregressive models used to model scalar time series. More precisely, we model a sequence of vectors in terms of an auto-regression: 
\begin{equation}\label{eq:VAR}
    x_{t+1} = W x_t + \varepsilon_t, \quad t=1, \cdots, T-1,
\end{equation}
where $W \in \RR^{d \times d}$ and $\varepsilon_t$ is a vector of white noise: $\EE(\varepsilon_t)=0$ and $\EE(\varepsilon_t \varepsilon^\top_{t'})=\Sigma$ if $t=t'$ and $0_{d\times d}$ otherwise, where $\Sigma$ is a symmetric positive definite matrix. When the noises $\varepsilon$ are Gaussian, assuming $\sum^{T-1}_{t=1}x_{t}x_t^\top$ is invertible, the maximum likelihood estimator of $W$ is given by $\hat{W} = [\sum^{T-1}_{t=1}x_{t+1}x_t^\top][\sum^{T-1}_{t=1}x_{t}x_t^\top]^{-1}$, and can therefore be estimated from ordinary least square regression, that is solving
\begin{equation}
\argmin_{W \in \RR^{d \times d}}\sum_{t=1}^{T-1} \|Wx_t - x_{t+1} \|^2.
\end{equation}

\paragraph{Least Square Flow.}
 Given a \textbf{fixed} sequence $x_{1:T}\eqdef(x_1, \cdots, x_{T}) \in \RR^{T \times d}$, we consider the least square optimization problem \eqref{eq:OLS}. A natural way of finding a solution is by gradient flow: $\dot{W_T}(s) = - \sum_{t=1}^{T-1}(W_T(s) x_t - x_{t+1})x_t^\top$.

Suppose we have an autoregressive model, trained to implement the least square flow: for a sequence $x_{1:\tmax}$, the model outputs for each $x_T$ the limit of the least square flow  $W_T(+\infty) = \hat{W}$, where $\hat{W} = [\sum^{T-1}_{t=1}x_{t+1}x_t^\top][\sum^{T-1}_{t=1}x_{t}x_t^\top]^{-1}$. An additional layer can simply output $\hat{W} x_T \simeq x_{T+1}$.
In addition, this proposed model is well suited for length generalization, that is the ability to extrapolate from short sequences to longer ones. Indeed, \mic{one should have $\hat{W} = W$ with probability one has soon as $T \geq d$.}
However, with such an approach, each token would have to carry the additional information $W_T(s)$ during the trajectory, which requires an additional $d^2$ space in memory for each token.  

We now present an alternate method based on duality to show that we can reduce this extra space to $d$. 
The idea is to consider a dual approach in order to replace each $W_T \in \RR^{d\times d}$ with a vector $\lambda_T \in \RR^{d}$. Instead of carrying the information $W_T(s)$ during the trajectory, one would simply need to store the information $\lambda_T(s)$.
We proceed as follows: the minimum Frobenius-norm solution of \eqref{eq:OLS}, \textit{when it exists}, satisfies:
\begin{equation}\label{eq:min_frob}
    W^* = \argmin_{A(W) = (x_2, \cdots x_{T})} \| W\|_F
\end{equation}
with $A(W) = (Wx_1, \cdots, Wx_{T-1}).$
The solution of \eqref{eq:min_frob} satisfies 
$W^* = A^\top(AA^\top)^{-1} x_{2:T},$
that is 
$W^* = A^\top \Lambda^*$ 
with 
$
\Lambda^* = (A A^\top)^{-1} x_{2:T}.
$ It turns out that one can compute the transpose of $A$ as follows.
\begin{prop}
    One has for all $\Lambda \eqdef (\lambda_1, \cdots, \lambda_T) \in \RR^{T \times d}$, $A^\top \Lambda = \sum_{t=1}^{T}\lambda_t x_t^\top.$
\end{prop}
\begin{proof}
    Indeed, one has 
    $\langleA(W), \Lambda\rangle_{\RR^{T \times d}} = \sum_{t, i}(Wx_t)_i\Lambda_{t,i} = \sum_{t,i,j}W_{i,j}x_{t,j}\Lambda_{t,i} = \sum_{ij}W_{ij} \sum_t{\Lambda_{t,i}x_{t,j}} = \langle W, \Bb \Lambda \rangle_{\RR^{d \times d}},$
with $\Bb \Lambda = \sum_{t=1}^{\tmax} \lambda_t x_t^\top$. By definition, one has $A^\top = \Bb$.
\end{proof}
Therefore, one could consider the dual problem 
\begin{equation}\label{eq:ls_dual}
\min_{\Lambda} \sum_{t=1}^{T-1} \| \sum_{\tau=1}^{T} \lambda_\tau x_\tau^\top x_{t} - x_{t+1}\|^2.
\end{equation}
This problem can be solved with gradient flow: 
\begin{equation}\label{eq:gf}
\dot{\Lambda} = -\nabla_{\Lambda}  \sum_{t=1}^{T-1} \| \sum_{\tau=1}^{T} \lambda_\tau x_\tau^\top x_{t} - x_{t+1}\|^2,
\end{equation}
There are two main issues with \eqref{eq:gf} in the context of autoregressive in-context learning in Transformers.
\begin{itemize}
    \item First, formulation \eqref{eq:gf} is not causal: the evolution of any $\lambda_t$ depends on the future $\lambda_\tau$ for $T \geq \tau \geq t$. Worse, different values of $T$ implies different dynamics for each $\lambda_t$.
    \item Second, the gradient in \eqref{eq:gf} involves several terms and cannot, to the best of our knowledge, be expressed as a linear attention layer. 
\end{itemize}
We therefore propose an alternate method to approach a solution of \eqref{eq:ls_dual} which will be causal and expressible with linear attention.
\paragraph{Modified Least Square Flow.}
We consider, for $1\leq T \leq \tmax$ the flow
\begin{equation}\label{eq:agf}
    \dot{\lambda}_T = - \nabla_{\lambda_T} \| \sum_{t=1}^T\lambda_tx^\top_tx_T - x_{T+1}\|^2 = - \|x_T\|^2 (\sum_{t=1}^T\lambda_tx^\top_tx_T - x_{T+1})
\end{equation}
To unsure the stability of the flow and to ease the expressions, we make the assumption that the vector autoregressive process has unit norm.
\begin{asp}\label{asp:unit}
    We have $\|x_T\| = 1$, $\forall T \geq 1$.  
\end{asp}
This assumption is easily satisfied when $x_1$ is of unit norm and $x_{T+1} = W x_T$ with $W$ orthogonal.
Under assumption \ref{asp:unit}, we can rewrite \eqref{eq:agf} as 
\begin{equation}
\label{eq:agf_unit}
    \dot{\Lambda} = - A \Lambda + B   
\end{equation}
with $A_{t,t'} = (x_{t'}^\top x_{t}) \times 1_{t'\leq t}$ a masked attention matrix and $B_t = x_{t+1}.$
At optimality, one necessarily has
$
    \sum_{t=1}^{T} \lambda_t x_t^\top x_T = x_{T+1}.
$
Let us define $W_T = \sum_{t=1}^{T} \lambda_t x_t^\top$. Then one has $W_T x_T = x_{T+1}$ for all $T$.
\begin{rem}
One has
$
\lambda_T = x_{T+1} - \sum_{t=1}^{T-1}\lambda_t x_t^\top x_T
$
so that the $\lambda_T$ themselves satisfy an autoregressive process.
\end{rem}
The build Transformer perfectly fits the training data: $W_T x_T = x_{T+1}$. A natural question concerns the length generalization properties of this model. In other words, does this modified least square flow leads to a correct estimation of the context matrix $W$? For instance, can we prove that if the $x_T$ are generated according to $x_{T+1} = Wx_T$, then 
$\| W_T - W\| \leq C(T)$
with $C(T)$ going to $0$ as $T \to +\infty$?
In order to show this result, we consider the following assumption which imposes to restrict the context matrices $W$ to orthogonal matrices, as done by \citet{vonoswald2023uncovering} and \citet{sander2024transformers}.
\begin{asp}
    Sequences are generated by drawing $W \sim \Uu(O(d))$ and $x \sim \Uu(S^{d-1})$, and $x_{T+1} = Wx_T$ for all $T$, with $x_1 \eqdef x$.
\end{asp}
Note that assumption \ref{asp:ortho} implies assumption \ref{asp:unit}. 
\begin{thm}[Length Generalization]
    Suppose assumption \ref{asp:ortho}. Then one has 
    $$
    W_{T+1} - W = - ( W (I - x x^\top))^{T+1} W^{-T}.
    $$

In addition, $\rho(W (I - x x^\top)) < 1$ with probability $1$ \mic{and 
    $$
    \EE_{W,x} \rho(W (I - x x^\top)) \leq (1 - \frac{1}{d})^{\frac{1}{2}}.
    $$}

    \mic{TODO: compare the tokens instead of the matrices}
\end{thm}

Therefore, the model achieves length generalization exponentially in the sequence length.

\paragraph{Highter order recursions.}

\paragraph{From Random Fourier Features to $2$-layer Perceptrons}

If $x_{t+1} = W \sigma (Ux_t)$, we get the flows for the minimization of $\frac12 \sum_{s=1}^t\|W\sigma(Ux_s) - x_{s+1}\|^2$
$$
\dot{W} = - \sum_{s=1}^t(W \sigma(Ux_s) - x_{s+1})\sigma(Ux_s)^\top
$$
and
$$
\dot{U} = - \sum_{s=1}^t\diag(\sigma'(Ux_s))W^\top(W \sigma(Ux_s) - x_{s+1})x_s^\top
$$
Inspired by these flows, we consider

$$
\dot{p_t} = - \sum_{s=1}^t(p_s - x_{s+1})\sigma(h_s)^\top \sigma(h_t)
$$
and
$$
\dot{h_t} = - \sum_{s=1}^t\diag(\sigma'(h_s))W^\top(W \sigma(h_s) - x_{s+1})x_s^\top x_t.
$$

\paragraph{Option 1.}

Form tokens $e_s = (u_s, x_s, x_{s+1})$. When we process a current token $e_t$ we mask the last coordinate so that $e_t$ becomes $(u_t, x_t, 0)$ and we can do $\dot{u}_t = -\sum_{s=1}^t \frac{k(x_s, x_t)}{\sum_{\tau=1}^t k(x_\tau, x_t)} (u_s - (e_s)_{2d:3d})$ with $(e_s)_{2d:3d} = 1_{s<t} x_{s+1}$. The problem is if think this cannot be done in parallel, one has to store the previous trajectory for the $u_s$, and the formulation for a token $e_t$ depends on whether it is the last token or not. Unless we can just mask the last coordinate of the current token just for computing the attention ? Would be something like $e_t = e_t + f(e_1, .., e_{t-1}, Me_t)$ for $e_t = (u_t, x_t, x_{t+1})$ and $Me_t = (u_t, x_t, 0)$. 

\paragraph{Option 2.}

Form the tokens $e_s = (x_{s-1}, x_s, u_s)$. In this case the difficulty comes from the fact that we would like to express $\frac{\sum_{s=1}^t k(x_{s-1}, x_t)}{\sum_{\tau=1}^{t+1}k(x_{\tau-1}, x_t)}x_s = \frac{\sum_{s=1}^t k(Qe_s, Kx_t)}{\sum_{\tau=1}^{t+1}k(Qe_{\tau}, K e_t)}Ve_s$ (with the convention $k(x_0, x) = 0$).

\paragraph{Option 3.} We consider true attentions and try to control the error term. We therefore implement: 

$$
\dot{u}_{t} = -\frac1{S_t(x_t)} \sum_{s=1}^tk(x_s,x_t)(u_s - 1_{s<t}x_{s+1}) + [\frac{1}{S_{t-1}(x_t)} - \frac{1}{S_{t}(x_t)}]\sum_{s=1}^tk(x_{s-1},x_t)x_s
$$
with $S_t(x) = \sum_{s=1}^t k(x_s, x)$.

At convergence one has $u = x_{2:T} - A^{-1}\diag(A)x_{2:T} + A^{-1}z$, where $z_t = [\frac{S_{t}(x_t)}{S_{t-1}(x_t)} - 1]\sum_{s=1}^tk(x_{s-1},x_t)x_s$. Writing $A \alpha = z$ we want to show that $\alpha_t \to 0$, where $\alpha$ satisfies:
$$
\sum_{s=1}^tk(x_s,x_t) \alpha_s = \frac{k(x_t, x_t)}{S_{t-1}(x_t)}\sum_{s=1}^tk(x_{s-1},x_t)x_s
$$
The right term can be seen as an expectation and is bounded. But there is no reason for the right term to converge right ?

\newpage

Suppose we have an autoregressive model, trained to implement the least square flow: for a sequence $x_{1:\tmax}$, the model outputs for each $x_T$ the limit of the least square flow  $W_T(+\infty) = \hat{W}$, where $\hat{W} = [\sum^{T-1}_{t=1}x_{t+1}x_t^\top][\sum^{T-1}_{t=1}x_{t}x_t^\top]^{-1}$. An additional layer can simply output $\hat{W} x_T \simeq x_{T+1}$.
In addition, this proposed model is well suited for length generalization, that is the ability to extrapolate from short sequences to longer ones. Indeed, \mic{one should have $\hat{W} = W$ with probability one has soon as $T \geq d$.}
However, with such an approach, each token would have to carry the additional information $W_T(s)$ during the trajectory, which requires an additional $d^2$ space in memory for each token.  

We now present an alternate method based on duality to show that we can reduce this extra space to $d$. 
The idea is to consider a dual approach in order to replace each $W_T \in \RR^{d\times d}$ with a vector $\lambda_T \in \RR^{d}$. Instead of carrying the information $W_T(s)$ during the trajectory, one would simply need to store the information $\lambda_T(s)$.
We proceed as follows: the minimum Frobenius-norm solution of \eqref{eq:OLS}, \textit{when it exists}, satisfies:
\begin{equation}\label{eq:min_frob}
    W^* = \argmin_{A(W) = (x_2, \cdots x_{T})} \| W\|_F
\end{equation}
with $A(W) = (Wx_1, \cdots, Wx_{T-1}).$
The solution of \eqref{eq:min_frob} satisfies 
$W^* = A^\top(AA^\top)^{-1} x_{2:T},$
that is 
$W^* = A^\top \Lambda^*$ 
with 
$
\Lambda^* = (A A^\top)^{-1} x_{2:T}.
$ It turns out that one can compute the transpose of $A$ as follows.
\begin{prop}
    One has for all $\Lambda \eqdef (\lambda_1, \cdots, \lambda_T) \in \RR^{T \times d}$, $A^\top \Lambda = \sum_{t=1}^{T}\lambda_t x_t^\top.$
\end{prop}
\begin{proof}
    Indeed, one has 
    $\langleA(W), \Lambda\rangle_{\RR^{T \times d}} = \sum_{t, i}(Wx_t)_i\Lambda_{t,i} = \sum_{t,i,j}W_{i,j}x_{t,j}\Lambda_{t,i} = \sum_{ij}W_{ij} \sum_t{\Lambda_{t,i}x_{t,j}} = \langle W, \Bb \Lambda \rangle_{\RR^{d \times d}},$
with $\Bb \Lambda = \sum_{t=1}^{\tmax} \lambda_t x_t^\top$. By definition, one has $A^\top = \Bb$.
\end{proof}
Therefore, one could consider the dual problem 
\begin{equation}\label{eq:ls_dual}
\min_{\Lambda} \sum_{t=1}^{T-1} \| \sum_{\tau=1}^{T} \lambda_\tau x_\tau^\top x_{t} - x_{t+1}\|^2.
\end{equation}
This problem can be solved with gradient flow: 
\begin{equation}\label{eq:gf}
\dot{\Lambda} = -\nabla_{\Lambda}  \sum_{t=1}^{T-1} \| \sum_{\tau=1}^{T} \lambda_\tau x_\tau^\top x_{t} - x_{t+1}\|^2,
\end{equation}
There are two main issues with \eqref{eq:gf} in the context of autoregressive in-context learning in Transformers.
\begin{itemize}
    \item First, formulation \eqref{eq:gf} is not causal: the evolution of any $\lambda_t$ depends on the future $\lambda_\tau$ for $T \geq \tau \geq t$. Worse, different values of $T$ implies different dynamics for each $\lambda_t$.
    \item Second, the gradient in \eqref{eq:gf} involves several terms and cannot, to the best of our knowledge, be expressed as a linear attention layer. 
\end{itemize}
We therefore propose an alternate method to approach a solution of \eqref{eq:ls_dual} which will be causal and expressible with linear attention.
\paragraph{Modified Least Square Flow.}
We consider, for $1\leq T \leq \tmax$ the flow
\begin{equation}\label{eq:agf}
    \dot{\lambda}_T = - \nabla_{\lambda_T} \| \sum_{t=1}^T\lambda_tx^\top_tx_T - x_{T+1}\|^2 = - \|x_T\|^2 (\sum_{t=1}^T\lambda_tx^\top_tx_T - x_{T+1})
\end{equation}
To unsure the stability of the flow and to ease the expressions, we make the assumption that the vector autoregressive process has unit norm.
\begin{asp}\label{asp:unit}
    We have $\|x_T\| = 1$, $\forall T \geq 1$.  
\end{asp}
This assumption is easily satisfied when $x_1$ is of unit norm and $x_{T+1} = W x_T$ with $W$ orthogonal.
Under assumption \ref{asp:unit}, we can rewrite \eqref{eq:agf} as 
\begin{equation}
\label{eq:agf_unit}
    \dot{\Lambda} = - A \Lambda + B   
\end{equation}
with $A_{t,t'} = (x_{t'}^\top x_{t}) \times 1_{t'\leq t}$ a masked attention matrix and $B_t = x_{t+1}.$
At optimality, one necessarily has
$
    \sum_{t=1}^{T} \lambda_t x_t^\top x_T = x_{T+1}.
$
Let us define $W_T = \sum_{t=1}^{T} \lambda_t x_t^\top$. Then one has $W_T x_T = x_{T+1}$ for all $T$.
\begin{rem}
One has
$
\lambda_T = x_{T+1} - \sum_{t=1}^{T-1}\lambda_t x_t^\top x_T
$
so that the $\lambda_T$ themselves satisfy an autoregressive process.
\end{rem}
The build Transformer perfectly fits the training data: $W_T x_T = x_{T+1}$. A natural question concerns the length generalization properties of this model. In other words, does this modified least square flow leads to a correct estimation of the context matrix $W$? For instance, can we prove that if the $x_T$ are generated according to $x_{T+1} = Wx_T$, then 
$\| W_T - W\| \leq C(T)$
with $C(T)$ going to $0$ as $T \to +\infty$?
In order to show this result, we consider the following assumption which imposes to restrict the context matrices $W$ to orthogonal matrices, as done by \citet{vonoswald2023uncovering} and \citet{sander2024transformers}.
\begin{asp}
    Sequences are generated by drawing $W \sim \Uu(O(d))$ and $x \sim \Uu(S^{d-1})$, and $x_{T+1} = Wx_T$ for all $T$, with $x_1 \eqdef x$.
\end{asp}
Note that assumption \ref{asp:ortho} implies assumption \ref{asp:unit}. 
\begin{thm}[Length Generalization]
    Suppose assumption \ref{asp:ortho}. Then one has 
    $$
    W_{T+1} - W = - ( W (I - x x^\top))^{T+1} W^{-T}.
    $$

In addition, $\rho(W (I - x x^\top)) < 1$ with probability $1$ \mic{and 
    $$
    \EE_{W,x} \rho(W (I - x x^\top)) \leq (1 - \frac{1}{d})^{\frac{1}{2}}.
    $$}

    \mic{TODO: compare the tokens instead of the matrices}
\end{thm}

Therefore, the model achieves length generalization exponentially in the sequence length.

\section{Non-Linear Case.}

We consider recursions of the form $x_{t+1} = f(x_t)$ with $f(x) \eqdef W \phi(x)$ where $\phi$ is the feature map associated with the un-normalized attention kernel : $\langle \phi(x) , \phi(y) \rangle \eqdef e^{\langle x , y \rangle} = k(x,y)$. One has $\phi(x) \in \Hh$ and $W : \Hh \to \RR^d$. 

\paragraph{Method 1.}

Unregularized kernel regression on $x_{1:T}$ amounts to finding $\alpha_{1:T}$ such that 

\begin{equation}\label{eq:unmasked}
\forall t \in [1, \cdots, T]: \quad \sum_{s=1}^T \alpha_s k(x_s, x_t) = f(x_t).
\end{equation}

Instead, we propose a masked method that amounts to finding $z_{1:T}$ such that 
\begin{equation}\label{eq:masked}
\forall t \in [1, \cdots, T]: \quad \sum_{s=1}^t z_s k(x_s, x_t) = f(x_t).
\end{equation}
We recall that this is done by having $z$ following the flow:

$$
\dot z = - A z + y
$$

with $A$ the masked attention matrix: $A_{t, s} = k(x_t, x_s) 1_{s \leq t}$ and $y_t = x_{t+1} = f(x_t)$.

Denoting $f_t(x) = \sum_{s=1}^t z_s k(x_s, x) = W_t \phi(x)$ with $W_t \eqdef \sum_{s=1}^t z_s \phi(x_s)^\top$, we can show that

$$
W_{t+1} - W = (W_t - W)(I - \frac{\phi(x_{t+1}) \phi(x_{t+1})^\top}{\phi(x_{t+1})^\top \phi(x_{t+1})}). 
$$

Note that this is a generalization of the method presented above in the linear case.

The convergence of $W_t$ to $W$ therefore depends on the convergence of the product of orthogonal projections $\Pi_{s=1}^T (I - \frac{\phi(x_{s}) \phi(x_{s})^\top}{\phi(x_{s})^\top \phi(x_{s})})$. We cannot hope to have this product going to zero in the general case, as any $y \in H$ which is orthogonal to all the $\phi(x_s)$ will stay the same under the action of this product. In any case, one cannot hope of having $W_t$ being a better estimate of $W$ than the one given by \eqref{eq:unmasked}. 
Looking at a prediction for a given $x \in \RR^d$, one has 

$$
f_{t+1}(x) - f(x) = f_t(x) - f(x) - \frac{k(x_{t+1}, x)}{k(x_{t+1}, x_{t+1})}(f_t(x_{t+1}) - f(x_{t+1})). 
$$

What choices of $x$ lead to this going to $0$ ? Can we show it at least for $x = x_{t+2}$ ?

One has $f_t(x_t) = x_{t+1}$ and $f_t(x_t) = f_{t-1}(x_t) + z_t k (x_t, x_t)$. Therefore, 
$$
f_{t-1}(x_t) = x_{t+1} + z_t e^{\|x_t\|^2}
$$
so we want to have $z_t \to 0$, which is weaker than having $W_t$ converging. 
\paragraph{Method 2.}

The second method consists in working directly on the predictions. We consider the flow 

\begin{equation}
    \dot u = - A u + (A - \diag (A))y
\end{equation}

with the same notations as for method 1. This can be expressed with an attention neural ODE. 

We want to show that $u^{\star}_t \eqdef u_t(+ \infty)$ is close to $y_t = x_{t+1}$. One has 

$$
u^{\star} = y - A^{-1} \diag (A) y.
$$

Writing $\lambda = A^{-1} \diag (A) y$, we obtain something similar to the recursions on $z$, that is 
$$
 \sum_{s=1}^t \lambda_s k(x_s, x_t) = k(x_t, x_t) f(x_t).
$$

We would like to show (under proper hypothesis) that $\lambda_t \to 0$ as $t \to + \infty$.

\paragraph{Reasoning for method 1.}

Theorem 1 of \url{https://www.math.uni.wroc.pl/~szwarc/ps/rainis.pdf} states that almost all linearly dense sequences $\{e_t\}$ of norm 1 in a Hilbert space will satisfy the property that the product of projections onto the orthogonal complement of $e_t$ $P_t \eqdef (I - e_t e_t^\top ) $ converges to 0:
$$
\lim_{t \to \infty} P_t P_{t-1} \cdots P_1(x) = 0 \text{ for all } x 
\text{ in a linearly dense subset of } H.
$$
We say that the sequence $(e_t)_t$ is effective.

This means that if we take an i.i.d. sequence $(e_t)_t$ with a distribution on the sphere $S(H)$ whose support is linearly dense in $H$, then it is almost surely effective. 

If we take $x_t$ i.i.d. according to a Gaussian distribution, then this induces a probability distribution on the sphere by considering the sequence defined by $e_t = \frac{\phi(x_t)}{\|\phi(x_t)\|}$. This distribution has a linearly dense support in $H$ by construction of $H$. Therefore, $(e_t)_t$ is almost surely effective.

Now if we take $e_t = \frac{\phi(x_t)}{\|\phi(x_t)\|}$ with i.i.d. $x_t$, I want to show that we can almost surely write $x_t$ as $x_{t+1} = W \phi(x_t)$ for some $W$. However, the $\phi(x_t)$ are almost surely linearly independent in H. This allows us to define $W$ as a linear map that sends $\phi(x_t)$ to $x_{t+1}$. 

So this means that the sequences $\frac{\phi(x_t)}{\|\phi(x_t)\|}$ for $x_t$ following a process $x_{t+1} = W \phi(x_t)$ have measure 1 in the set of sequences of the form $e_t = \frac{\phi(x_t)}{\|\phi(x_t)\|}$ which themselves are almost surely effective.

\newpage

\section{Traditional In-Context Learning with Kernel Attention descent}

We consider $x_{i} \in \mathbb{R}^d$ be the covariates drawn i.i.d. from a distribution and $f \in \Hh$. We define $y_i = f(x_i) \in \RR$ for $1 \leq i \leq n$ (easily extendable to $\RR^d$). We consider the input sequence 
$$
e = 
\left[ \begin{array}{ccccc}
x_{1} & x_{2} & \cdots & x_{n} & x_{n+1} \\
y_{1} & y_{2} & \cdots & y_{n} & 0 \\
z^0_{1} & z^0_{2} & \cdots & z^0_{n} & 0 \\
\end{array} \right] \in \mathbb{R}^{(d+2)\times(n+1)},
$$
As is standard, we place a zero in place of the prediction $\hat{y}_{n+1}$, that is we take $y_{n+1} = 0$. We also take $z^0_{n+1} = 0$. The goal is to build a Transformer that will predict $f(x_{n+1})$. For this, we consider the neural ODE on $z \eqdef (z_1, \cdots, z_{n+1})$:

$$
\dot z = - A z + y,
$$

where $A$ is the attention matrix defined as $A_{i,j} = 1_{j < n+1}k(x_i, x_j)$. Note that we mask the last column as is standard \citep{ahn2023transformers}. Then at convergence of the flow one has: 
$$
\forall 1 \leq i \leq n : \quad \sum_{j=1}^n k(x_i, x_j) z_j = y_i. 
$$

It then suffices to add a last module which outputs $\sum_{j=1}^{n} k(x_{n+1}, x_j) z_j$ to obtain an estimate of $f(x_{n+1})$.

\subsection{Particular case.}

We assume $x_{t+1} = W x_t$ but still use a non-linear kernel. We want to use the result on the measures. 

We have 
$$
e^{\cos(x)} = \sum_{k=-\infty}^{\infty} I_k(1) e^{ikx}
$$

where $I_k$ is the modified Bessel function. The above identity comes from the fact (\url{https://appliedmath.brown.edu/sites/default/files/fractional/35%

$$e^{z\left(t + \frac{1}{t}\right)/2} = \sum_{k=-\infty}^{+\infty} I_k(z) t^k, \quad 0 < |t| < \infty,$$

where we take $t = e^{ix}$ and $z=1$.

Then, for $ \mu = \sum_{k=-\infty}^{\infty} \frac{I_k(1)}{e} \delta_{k \theta}$, we have 

$$
\int e^{in\phi} d\mu(\phi) = e^{\cos{n \theta}- 1}.
$$

This solves the $2d$ case. Now for the general case of $W$ orthogonal matrix, up to decomposing into rotations, there exists $\theta_i$'s such that $a_n \eqdef k(x, x_n) = \exp{(\langle x, W^n x \rangle)} = g(n(\theta_1, \cdots, \theta_p))$ where $g$ is a $p$ dimensional periodic function= $g(x + (T, \cdots, T)) = g(x)$. Then $g$ admits a Fourier series and one has 
$$
g(x) = \sum_{k \in \ZZ^p} c_k e^{i\langle x, k \rangle}.
$$
Therefore, 
$a_n = \sum_{k \in \ZZ^p} c_k e^{i n \langle \theta, k \rangle},$ and we take 

$$
\mu = \sum_{k \in \ZZ^p} c_k \delta_{\langle \theta, k \rangle}.
$$

\begin{proofsketch}
    We concatenate positional encodings $ (p_0, p_1, \cdots, p_T) \in \mathbb{R}^{T+1} $ to $ x_{0:T} $ and obtain a new sequence $ x^p_{0:T} $ with $ x^p_t = (x_t, p_t) \in \mathbb{R}^{d+1} $. We explicitly take $ p_t = (-1)^{t} \alpha t $ for some $ \alpha > 0 $. We consider a two-head attention module with no residual connection, corresponding to an attention cost:
    $
    \mathcal{A}^h_{t,s} = \frac{e^{(-1)^h p_s p_t}}{\sum_{\tau=1}^{t} e^{ (-1)^h p_{\tau} p_t}}.
    $
    The forward rule is then defined as, for $ t \geq 1 $:
    $
    T^{\alpha}(x_t) = \sum_{s=1}^t \mathcal{A}^1_{t,s} \cdot (x_s, 0, 0) + \sum_{s=1}^t \mathcal{A}^2_{t,s} \cdot (0, x_s, 0).
    $
    As $ \alpha \to +\infty $, one has $ T^{\alpha}(x_t) \to (x_{t-1}, x_t, 0) $. 
\end{proofsketch}

Consider the augmented tokens $e_t = (0, x_{t+1}, x_{t}).$ They can be formed using a one layer attention with $2$ heads and positional encoding:
\begin{lem}\label{lem:augmented_heads}
The tokens $e_{1:T}$ can be approximated with arbitrary precision given tokens $x_{1:T}$ with a positional encoding-only Transformer. 
\end{lem}
\begin{proof}
Let, for $x \in \RR^d$, $W^1_V x \eqdef (0, x, 0) \in \RR^{3d}$ and $W^2_V x \eqdef (0, 0, x)$. 
We now simply consider for $h \in \{1,2\}$: $$\mathcal{A}^{h}_{t,:}=\text{softmax}(P^h_{t, :}).$$
We choose $P^1$ and $P^2$ such that ${A}^{1}_{t,t'} \simeq \delta_{t'=t+1}$ and ${A}^{2}_{t,t'} \approx \delta_{t'=t}.$
Then
$$
\sum_{h=1}^{2} \sum_{t'=1}^{t} \mathcal{A}^h_{t,t'} W^h_V x_{t'} \simeq (0, x_{t+1}, x_{t}) = e_t. 
$$
\end{proof}
We therefore, instead of considering the tokens $(x_1, \cdots x_{\tmax +1})$, consider $(e_1, \cdots e_{\tmax})$.

Now, given tokens $e_{1:\tmax}(s)$ with $e_t(s) = (\lambda_t(s), x_{t+1}, x_t)$ and $\lambda_t(0) = 0$, we can form 
$$-\sum_{t=1}^{T} \lambda_t(s) x_t^\top x_T$$
with a linear attention, by considering
$$Att(e_t(s), e_t'(s)) = x_t^\top x_{t'}$$ 
and $W_V e_t(s) = (-\lambda_t(s), 0, 0).$ Indeed, the output of $e_T$ by such a layer is precisely 
$$
\sum_{t=1}^T Att(e_T(s), e_t(s)) W_Ve_t(s) = (-\sum_{t=1}^{T} \lambda_t(s) x_t^\top x_T, 0, 0).
$$
A second head can simply form $e_T \to (x_{T+1}, 0, 0)$ using positional encoding, similarly to Lemma \ref{lem:augmented_heads}.

In other words, one can compute  the right hand side of \eqref{eq:agf_unit} given the tokens $e_t$ with a $2$ heads transformer, for which the associated autoregressive Neural ODE writes 
$$
\frac{de}{d\tau}_T = (-\sum_{t=1}^{T} \lambda_t(s) x_t^\top x_T + x_{T+1}, 0, 0), 
$$
which is equivalent to \eqref{eq:agf_unit}, and $\dot{x}_T = 0$ $\forall T$. Finally, one last layer can simply implement $\sum_{t=1}^{T} \lambda_t(+\infty) x_t^\top x_{T+1} $.

We have 
$$
e^{\cos(x)} = \sum_{k=-\infty}^{\infty} I_k(1) e^{ikx}
$$

where $I_k$ is the modified Bessel function. The above identity comes from the fact (\url{https://appliedmath.brown.edu/sites/default/files/fractional/35%

$$e^{z\left(t + \frac{1}{t}\right)/2} = \sum_{k=-\infty}^{+\infty} I_k(z) t^k, \quad 0 < |t| < \infty,$$

where we take $t = e^{ix}$ and $z=1$.

Then, for $ \mu = \sum_{k=-\infty}^{\infty} \frac{I_k(1)}{e} \delta_{k \theta}$, we have 

$$
\int e^{in\phi} d\mu(\phi) = e^{\cos{n \theta}- 1}.
$$

This solves the $2d$ case. Now f

\end{document}